\documentclass[runningheads]{llncs}

\usepackage{eccv}

\usepackage{eccvabbrv}

\usepackage{graphicx}
\usepackage{booktabs}
\usepackage{fontawesome}

\usepackage[accsupp]{axessibility}  %

\usepackage[]{tabularx}
\usepackage{multirow}
\usepackage[title]{appendix}
\usepackage{pgfplots}

\usepackage{soul}
\usepackage{xcolor,colortbl}
\usepackage{graphicx}
\usepackage{subcaption}
\usepackage{tikz}
\usetikzlibrary{arrows,calc,spy, babel}

\usepackage{amsmath}

\usepackage{etoolbox}   %
\providetoggle{arxiv}
\settoggle{arxiv}{false}

\newcommand*\samethanks[1][\value{footnote}]{\footnotemark[#1]}

\newcommand{\goldmedal}{{\color{orange}\faTrophy}}
\newcommand{\silvermedal}{{\color{gray}\faTrophy}}

\usepackage[breaklinks,colorlinks]{hyperref}

\usepackage{orcidlink}

\begin{document}
\settoggle{arxiv}{true}

\title{MeshFeat: Multi-Resolution Features\\for Neural Fields on Meshes}

\author{
Mihir Mahajan\thanks{Equal contribution}\inst{1} %
\and
Florian Hofherr\samethanks\inst{1,2} %
\and
Daniel Cremers\inst{1,2} %
}

\authorrunning{M.~Mahajan et al.}

\institute{
Technical University of Munich \and
Munich Center for Machine Learning
}

\maketitle

\begin{abstract}
Parametric feature grid encodings have gained significant attention as an encoding approach for neural fields since they allow for much smaller MLPs, which significantly decreases the inference time of the models. In this work, we propose MeshFeat, a parametric feature encoding tailored to meshes, for which we adapt the idea of multi-resolution feature grids from Euclidean space. We start from the structure provided by the given vertex topology and use a mesh simplification algorithm to construct a multi-resolution feature representation directly on the mesh. The approach allows the usage of small MLPs for neural fields on meshes, and we show a significant speed-up compared to previous representations while maintaining comparable reconstruction quality for texture reconstruction and BRDF representation. Given its intrinsic coupling to the vertices, the method is particularly well-suited for representations on deforming meshes, making it a good fit for object animation.
  
\end{abstract}
\section{Introduction}
\label{sec:intro}
\newcommand{\teaserfactor}{0.1}

\newcommand{\redcol}{red!50}
\newcommand{\greencol}{blue!50}
\newcommand{\bluecol}{yellow!75}

\newcommand{\puttri}[1]{
    \fill [\redcol] ([xshift=2pt,yshift=10pt]#1.center) circle [radius=1pt];
    \fill [\bluecol] ([xshift=5pt,yshift=8pt]#1.center) circle [radius=1pt];
    \fill [\greencol] ([xshift=5pt,yshift=12pt]#1.center) circle [radius=1pt];
}

\usetikzlibrary{positioning, calc}

\begin{figure}
    \centering

    \includegraphics[width=1.0\textwidth]{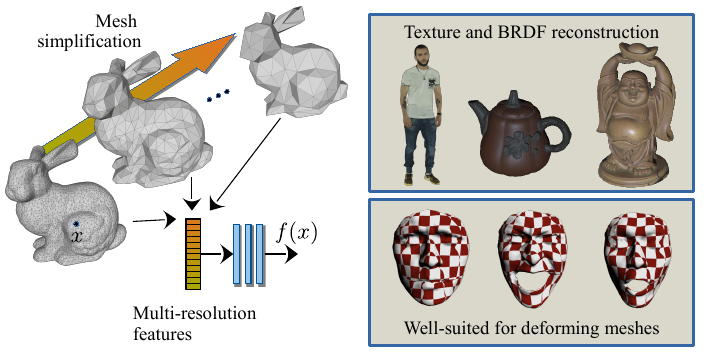}

    \caption{We present MeshFeat, a parametric encoding strategy for neural fields on meshes. We propose a multi-resolution strategy based on mesh simplification to enhance the efficiency of the encoding. Our approach allows for much smaller MLPs than previous frequency-based encodings, resulting in significantly faster inference times. We evaluate the method for texture reconstruction and BRDF estimation and demonstrate, that the encoding is well suited to represent signals on deforming meshes.}
    \label{fig:teaser}
\end{figure}

Capturing and modeling our 3D world realistically and effectively is becoming increasingly relevant
for AR, VR, CGI, simulations, and computer games. 
Classical approaches for scene representation often rely on explicit discrete data structures like voxel grids to store function values like SDF. Alternatively, point clouds, and meshes are used to store geometry or UV-Maps, and discrete texture maps to represent appearance. For all these approaches, the memory footprint is strongly coupled to the resolution, making high-resolution representations often impractical.
Since the breakthrough work by Mildenhall et al. \cite{mildenhall2020nerf}, the focus has shifted towards neural implicit representations, 
which use Multilayer Perceptrons (MLP) to represent scenes in a continuous and resolution-free manner.

The key to high-quality reconstructions is to use input encodings to overcome the spectral bias of neural networks \cite{rahaman2019spectral} and enable the model to learn high-frequency details. Models based on frequency encodings like positional encoding \cite{transformer, mildenhall2020nerf} or Fourier features \cite{rahimi2007random, tancik2020fourier} tend to be costly and slow to evaluate since all the information on the scene is stored in the weights of the MLP. This results in a strong coupling of different parts of the scene and necessitates large neural networks.
On the other hand, the use of multi-resolution feature grids \cite{DeepLocalShapes, liu2020neural, Local_Implicit_Grid_CVPR20, mehta2021modulated, PengNMP020ConvOccNets, SunSC22, 47782, yu2022plenoxels,li2023neuralangelo} as input encoding has led to a large gain in evaluation speed, making the models real-time capable. The idea of feature grids is to store local information about the scene in spatially distributed features and use only a small MLP to decode this information to the quantity of interest. This decoupling of spatial information from information decoding, combined with the resulting possibility of smaller networks, is the reason for the performance gain for those approaches. 

While neural representations have demonstrated impressive performance, numerous 3D computer graphics pipelines continue to utilize meshes as a fundamental data structure. Meshes remain a preferred choice in many applications due to the existence of efficient algorithms, intuitive editing capabilities, and convincing animation possibilities. Consequently, a noteworthy avenue of research explores the integration of mesh-based representations with neural fields, aiming to capitalize on the strengths of both approaches.
Texture fields by Oechsle \etal \cite{Oechsle19TextureFields} 
embed the points on the mesh into the surrounding Euclidean space and
leverages neural fields to regress texture values.
NeuTex \cite{Xiang21NeuTex} and other methods \cite{baatz2022nerftex, tancik2020fourier} follow the same idea, but additionally use frequency encodings to capture higher frequency details and circumvent the spectral bias issue. 
Unfortunately, these Euclidean embeddings do not take the mesh geometry into account, and for points where Euclidean and geodesic distance varies significantly, these methods can show bleeding artifacts \cite{Koestler22IntrinsicNeuralFields}.
Intrinsic neural fields by Koestler \etal \cite{Koestler22IntrinsicNeuralFields} addresses this by transferring the idea of frequency encodings from the Euclidean space to manifolds by using the eigenfunctions of the Laplace-Beltrami Operator. This makes the encoding intrinsic to the mesh. 
While they show promising results in reconstruction quality, they require heavy preprocessing to calculate the eigenfunctions.
Moreover, they require a large number of eigenfunctions, resulting in a substantial memory footprint of the model. All methods employ frequency-based encodings and, therefore, suffer from the requirement of large networks, hindering efficient inference and rendering. 

In this work, we propose MeshFeat, a novel parametric multi-resolution feature encoding for neural fields on meshes, for which we adopt ideas of feature grids from Euclidean space. We leverage the spatial structure given by the mesh and employ a mesh simplification algorithm for our multi-resolution strategy. Similarly to feature grids in Euclidean space, decoupling spatial information and decoding allows for much smaller MLPs and significantly accelerates the inference compared to previous methods. A schematic overview of our approach can be seen in \cref{fig:teaser}.

We summarize our main contributions as follows:

\begin{itemize}
\item We introduce MeshFeat, a parametric feature encoding for neural fields on meshes which includes an effective multi-resolution strategy based on mesh simplification.
\item We demonstrate that this feature encoding allows for significantly smaller MLPs compared to previous approaches, leading to a large speed-up in inference time.
\item We perform a thorough comparison of our approach to state-of-the-art methods for the tasks of single-object texture reconstruction and BRDF estimation. Moreover, we demonstrate the native applicability of our approach to signal representation on deforming meshes.
  
\end{itemize}

\noindent
Please see our project page at 
\url{https://maharajamihir.github.io/MeshFeat/}
for the source code.

\section{Related Work}
\label{sec:relatedwork}

\subsubsection{Neural Fields}
Scene representation based on coordinate-based networks has been used extensively in recent years. The idea is to use a simple multi-layer perception (MLP) to map a spatial position to quantities of interest like occupancy \cite{Occupancy_Networks,ChenZ19,PengNMP020ConvOccNets}, signed distance \cite{DeepSDF,GroppYHAL20IGR}, texture \cite{Oechsle19TextureFields}, density and color for volume rendering \cite{mildenhall2020nerf,YarivGKL21VolSDF,WangLLTKW21Neus} and more. In contrast to classical methods for scene representation purely based on explicit and discrete data structures like voxel grids,
these implicit approaches are not limited by a discrete resolution resulting in a much more memory-efficient representation.
The key to high-fidelity reconstructions with neural fields is to encode the input to a higher dimensional space to overcome the low-frequency bias of neural networks \cite{rahaman2019spectral,BasriGGJKK20FrequencyBiasInNeuralNetworks}.

\subsubsection{Input Encodings}
The different input encoding strategies can be divided into two categories: frequency-based encodings and parametric approaches. Among the most popular choices for the first type are positional encodings, which originate from transformer architectures \cite{transformer} and have been first introduced in the seminal NeRF paper \cite{mildenhall2020nerf} for neural fields. The idea is to use a sequence of sine and cosine functions to encode the input coordinates. Tancik \etal analyze these positional encodings using neural tangent kernel analysis and generalize the idea to the Fourier feature encoding \cite{tancik2020fourier}. A similar effect is achieved by SIRENs, which are MLP with periodic activation functions \cite{Sitzmann20SIRENS}. Hertz \etal complement this idea with an adaptive frequency masking scheme \cite{Hertz21SaAPE}.

In contrast, parametric encodings do not directly apply an encoding function to the input. Instead, they use the input coordinate as an interpolation point for learnable features stored in more classical data structures like grids \cite{DeepLocalShapes, liu2020neural, Local_Implicit_Grid_CVPR20, mehta2021modulated, PengNMP020ConvOccNets, SunSC22, 47782, yu2022plenoxels,li2023neuralangelo}. To evaluate the model, the grid features are interpolated at the input coordinate, and the result is used as an input for the MLP. 
Extending a single-resolution grid to a multi-resolution approach adds different scales of locality to the model. This decoupling of local and global information helps to improve efficiency and convergence properties. As further improvements, the use of hash grids \cite{mueller2022instant} and octrees \cite{takikawa2021nglod} have been proposed.

Parametric encodings allow for much smaller MLPs than their frequency-based counterparts, making evaluating these models significantly faster. Intuitively, 
feature encodings contain
the information ``What is at this position'' and the MLP decodes this 
into quantities of interest. In contrast, frequency-based encodings do not contain learned information, and the MLP needs to store all the information on the scene in its weights. The smaller computational cost for parametric encodings comes at the price of a larger memory footprint.

\subsubsection{Input Encodings for Neural Fields on Meshes}
Encoding strategies specifically for meshes 
have
received much less attention than encodings for neural fields in the Euclidean space. A common strategy, used \eg by Texture Fields \cite{Oechsle19TextureFields}, is to embed the points on the mesh into the surrounding Euclidean space.
This enables the use of established encoding strategies like Fourier features, as done by Text2Mesh \cite{Michel22Text2Mesh} or variants of positional encoding, as done by Hertz \etal \cite{HertzPGSC23MeshDraping}.

The drawback of this strategy is that the embedding to the surrounding space makes the encodings \emph{extrinsic} and the properties of the underlying manifold represented by the mesh are not considered, which can lead to artifacts \cite{Koestler22IntrinsicNeuralFields}. Koestler \etal show, that the equivalent to positional encoding in the Euclidean space 
are
the eigenfunctions of the Laplace-Beltrami Operator on manifolds \cite{Koestler22IntrinsicNeuralFields}. Grattarola and Vandergheynst investigate a similar idea \cite{Grattarola22GeneralisedImplicitNeuralRepresentations}. This results in an \emph{intrinsic} encoding that respects the properties of the manifold and enables applications like texture transfer. This approach has been used successfully to model deformation fields on meshes \cite{Lin23LevaragingIntrinsicPropertiesForNonRigidGarmentAlignment, Walker23ExplicitNeuralSurfaces} and for scene stylization \cite{Hwang23Text2Scene}.
While 
intrinsic encodings are elegant, in practice, many eigenfunctions have to be stored per vertex, resulting in a large model size. 
Also,
the computation of the eigenfunctions is costly since the eigenvalues of a large matrix need to be computed. 

All 
approaches for encodings on meshes discussed so far follow the idea of frequency encodings and 
thus
require large MLPs for high-quality reconstruction, resulting in slow evaluation times. In contrast, we present a parametric encoding on a mesh, 
enhanced by
a multi-resolution strategy. Only very few previous works consider this form of encoding.
Yang \etal \cite{YangBZBZCZ22NeuMesh} train features on a mesh, which they use as a scaffold for a neural scene representation in 3D. In contrast to our work, their approach does not contain hierarchical features, and moreover, they do not consider a signal restricted to the mesh surface. While Kim \etal propose multi-resolution features, they only consider spherical meshes \cite{Kim24HybridNeuralRepresentationsForSphericalData}. In contrast, we are not restricted to a certain class of meshes.

\subsubsection{Multi-Resolution Approaches on Meshes}
Multi-resolution approaches for meshes have been used for various tasks.
Lee \etal use mesh simplification based on vertex removal for adaptive re-meshing \cite{Lee98MAPS}.
Liu \etal investigate mesh simplification for multi-grid solvers on curved surfaces \cite{Liu21SurfaceMultigridViaIntrinsicProlongation}.
Jiang \etal propose a prismatic shell for meshes that can help to avoid geometric artifacts for mesh simplification algorithms for extreme vertex reductions at the cost of additional computational overhead \cite{Jiang20BijectiveProjectionInAShell}.
MeshCNN uses a learned mesh-pooling for a multi-resolution approach to mesh analysis tasks \cite{Hanocka19MeshCNN}.

\section{Method}
\label{sec:method}
In this section we present MeshFeat, a parametric encoding strategy for the parametrization of neural fields directly on meshes.
We adapt the idea of multi-resolution grids for the Euclidean space to meshes to obtain an efficient encoding. Instead of using a regular voxel grid in Euclidean space, we use the mesh vertices as pre-defined locations to store the feature vectors. We apply a mesh simplification algorithm to obtain different resolutions of the initial mesh for our multi-resolution approach. Similar to feature grids in Euclidean space, our approach allows to use very small MLPs resulting in a fast evaluation time. 

\subsection{Multi-Resolution Feature Encoding}
\newcommand{\imgwidth}{0.225}
\newcommand{\arrowheight}{0.175}

\begin{figure}[ht]
    \centering

    \includegraphics[width=1.0\textwidth]{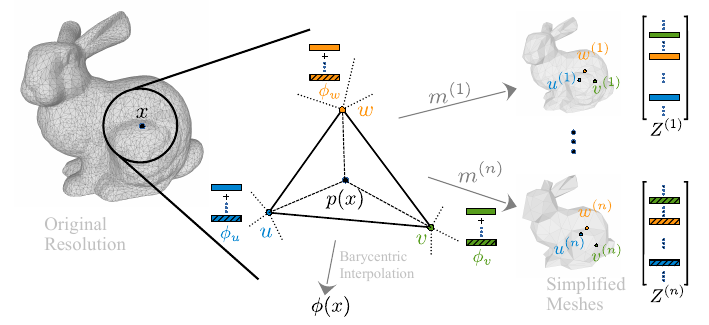}

\caption{
    Overview of our multi-resolution feature approach on the mesh. To get the feature encoding $\phi(x)$ for a point $x$ on the original mesh, we determine the vertices $u, v, w$ of the respective triangle. Using the mappings $m^{(i)}$, we gather the corresponding features from the different resolutions. By summing them, we obtain the features $\phi_u, \phi_v, \phi_w$ at the vertices in the original mesh. We receive the final feature encoding $\phi(x)$ by barycentric interpolation of the features at the vertices.
}
    \label{fig:architecture}
\end{figure}
\subsubsection{Mesh Simplification}
To obtain our multi-resolution feature encoding, we simplify the initial mesh $M = (V, F)$ to multiple resolutions using the quadric error metric decimation by Garland and Heckbert \cite{Garland98SimplifyingSurfacesWithColorAndTexture,GarlandH97SurfaceSimplification}. We denote the sequence of resolutions for the mesh simplification by $(r^{(i)})_i$, where $r^{(i)}\in[0,1]$ means, that the $i$-th resolution has $|V^{(i)}| = r^{(i)}|V|$ number of vertices. This yields the sequence of meshes $((V^{(i)}, F^{(i)}))_i$. Moreover, we store the mapping 
\begin{equation}
    m^{(i)}: V\rightarrow V^{(i)}\quad v\mapsto m^{(i)}(v)=v^{(i)}
    \label{eq:multi-res-mapping}
\end{equation}
that assigns a vertex $v\in V$ in the original mesh to the vertex $v^{(i)}\in V^{(i)}$ in resolution $i$ to which $v$ was collapsed in the decimation algorithm. Note that in contrast to other simplification-based multi-resolution schemes, we do not compute a geometric mapping between the resolutions, which reduces the computational overhead.
Please see the appendix for further information.

\subsubsection{Multi-Resolution Strategy}
For each of the resolutions, we use a learnable feature matrix $Z^{(i)}\in\mathbb{R}^{|V^{(i)}|\times d}$, where $d\in\mathbb{N}$ is the feature dimension. We denote the feature vector of vertex $v^{(i)}\in V^{(i)}$ by $Z^{(i)}(v^{(i)})\in\mathbb{R}^d$.

In contrast to parametric encodings on Euclidean grids, we do not interpolate the features for each resolution but rather accumulate the features from all resolutions on the original resolution and interpolate the results only once. The reason is that a point on the mesh for a certain resolution does not have a direct meaning on the meshes of the other resolutions. The Euclidean embedding of the point, for example, does not even need to be on the other meshes. Therefore, we use the mappings $m^{(i)}$ to ``pull'' all features of the coarser resolutions to the finest resolution (which is where we want to evaluate the function on the mesh). 
We obtain the combined feature vector $\phi_v$ for a vertex $v\in V$ in the original mesh by summing the feature vectors from all resolutions
\begin{equation} \label{eq:feature_acc}
\phi_v =  \sum_i Z^{(i)}(m^{(i)}(v)).
\end{equation}
See \cref{fig:architecture} for an overview of our multi-resolution feature pipeline.

\subsubsection{Feature Interpolation}
To obtain the feature vector for an arbitrary point $x$ on the mesh, we compute the barycentric coordinates $p(x)=[\lambda_1, \lambda_2, \lambda_3]^\top$ within the respective triangle of the original mesh. Let $v_1, v_2, v_3\in V$ be the vertices of this triangle. Then the encoding $\phi(x)$ for the point $x$ reads
\begin{equation}
    \label{eq:bary_interp}
    \phi(x) = \sum_i\lambda_i\phi_{v_i} = [\phi_{v_1}, \phi_{v_2}, \phi_{v_3}]~p(x).
\end{equation}

\subsection{Feature Regularization}
Depending on the training setup, it can happen that the features for some vertices will never receive a training signal. For example, for training from images and very fine meshes, some triangles might not be hit during the ray mesh intersection leading to missing supervision for vertices in that region.
We use a regularization of the features based on the mesh Laplacian to avoid the resulting artifacts. We denote the mesh Laplacian of the mesh in the original resolution by $L\in\mathbb{R}^{|V|\times |V|}$. To remove the dependency on the mesh scale and avoid the necessity to tune the regularization weight to each mesh individually, we normalize $L$ by its spectral norm, \ie we consider $\hat{L}=L/\|L\|_2$. For the regularization, we accumulate the features for all vertices of the original resolution in the matrix $\Phi\in\mathbb{R}^{|V|\times d}$ and sum the absolute values of the normalized Laplacian applied to the features, \ie

\begin{equation}\label{eq:reg_laplacian}
    \mathcal{L}_{\text{reg}} = \sum_{i,j}|(\hat{L}\Phi)_{i,j}|.
\end{equation}
This can be seen as the sum over the 1-norms of the Laplacian applied to the individual entries of the latent codes. This loss term penalizes large variations of neighboring feature values since the Laplacian computes the deviation from the local average. Still,
due to
the 1-norm, we allow for sparse larger changes. Note that while we compute the loss term only for the original resolution, the regularization 
affects all resolutions since we apply the Laplacian to the accumulated features. In practice, we use the robust Laplacian by Sharp and Crane \cite{Sharp20ALaplacianForNonmanifoldTriangleMeshes}.

\subsection{Model Architecture and Training Details}
To decode the feature vectors into the final function value, we employ a standard MLP with ReLU activation functions. For the experiments in this paper, we use a hidden dimension of 32 with 2 hidden layers. Also, we employ a sigmoid output-nonlinearity since all of our reconstructed signals are in the range $[0,1]$. To prevent overfitting, a weak L2 regularization (with weight $10^{-5}$) proved to be helpful.

We found that a total of 4 resolutions $r^{(i)} \in \{1, 0.1, 0.05, 0.01\}$ for the mesh simplification achieved optimal results for all experiments. Note, that for $r^{(1)} = 1$, the mapping $m^{(1)}$ yields the identity function, \ie we use the original mesh as our finest resolution.
We initialize the feature matrices using a normal distribution with $\sigma = 5\cdot 10^{-4}$. 

For training, we use a batch size of $b=8000$. 
We train for 1000 epochs for the texture reconstruction task and for 500 epochs for the BRDF estimation. We found that different learning rates for the MLP parameters and the features are crucial. 
We use $\text{lr}_\theta=2\cdot10^{-4}$ for the weights of the neural network and $\text{lr}_Z=5\cdot10^{-3}$ for our latent codes.
We use a factor $\lambda_{\text{reg}}=1.5\cdot10^{-6}$ to balance the regularization loss with the data loss.

\section{Experiments}
\label{sec:experiments}

In this section, we present a detailed evaluation of our proposed approach. First, we evaluate our method for single-object texture reconstruction from multi-view images. We compare against state-of-the-art methods \cite{Koestler22IntrinsicNeuralFields,Oechsle19TextureFields, Xiang21NeuTex}. Moreover, we perform an ablation study to evaluate the significance of our modeling choices.
As a second application of our method, we consider the reconstruction of a parametric BRDF from multi-view images of a single object with calibrated lighting. We use our approach to estimate the parameters of the well-known Disney BRDF \cite{burley2012physically} spatially varying on the mesh.
Furthermore, we demonstrate that our method is well-suited to represent quantities on deforming meshes due to the tight coupling with the mesh.
We refer to the supplementary material for additional experiments and further training information.

\subsubsection*{Concurrent Models}

We compare our encoding strategy to state-of-the-art work for neural fields on meshes with different encoding strategies. We use Texture Fields (TF) \cite{Oechsle19TextureFields} as a method with an \emph{extrinsic} encoding strategy. Consistent with the experiments done by Koestler \etal \cite{Koestler22IntrinsicNeuralFields}, we augment the method with random Fourier features (RFF). Moreover, we compare against Intrinsic Neural Fields (INF) \cite{Koestler22IntrinsicNeuralFields}, which uses an intrinsic and frequency-based encoding based on the eigenfunctions of the Laplace-Beltrami operator. Both methods use an MLP with 6 hidden layers of width 128.
To keep the experiments consistent with previous work \cite{Koestler22IntrinsicNeuralFields}, we also include a modified version of NeuTex \cite{Xiang21NeuTex} in our experiments on texture reconstruction, even though the method was originally designed for geometry estimation along with texture. NeuTex follows a UV-mapping-based approach and learns separate networks for geometry, UV-mapping and texture. Following Koestler \etal \cite{Koestler22IntrinsicNeuralFields}, we allow it to take advantage of the given geometry. We refer the reader to their work for details on this modification. 
Finally, as a reference, we include a non-neural baseline, for which we learn color values directly on the vertices of the original mesh and use barycentric interpolation to obtain the values on the triangles. 
Again, we use our regularization term to account for unsupervised vertices.
All methods are adapted into the same pipeline for a fair comparison.

\subsection{Texture Reconstruction from Multi-View Images}
\label{sec:TexRecon}

    \begin{table*}[t]  %
    \centering  %
    \begin{tabular}{ | l || l | l l l l l|} 
    \hline
    Object & Method & PSNR $\uparrow$ & DSSIM $\downarrow$ & LPIPS $\downarrow$ & \# Params. 
    $\downarrow$ & Speedup $\uparrow$  \\	\hline
\hline 
	\multirow{1}{5em}{human \newline $|V| = 129$k}
 
    & NeuTex \cite{Xiang21NeuTex} & 27.32 & 0.549 & 0.954 & 793k & 1.0x \\ 
    & TF+RFF \cite{Oechsle19TextureFields, tancik2020fourier} & 32.10 & 0.232 & 0.423 & \textbf{331k} \goldmedal & 1.96x \\ 
    & INF  \cite{Koestler22IntrinsicNeuralFields} & 32.46  \silvermedal & 0.215 & \textbf{0.390} \goldmedal & 133130k & 3.06x \silvermedal \\\cline{2-7}

    & Ours ($\lambda_{reg}=0$) & 31.25 & 0.323 & 0.510 & 604k & 13.49x \\ 
    & Ours ($d\uparrow$) & 32.46 & 0.203  \silvermedal & 0.410 & 1058k & 12.08x \\
    & Ours ($d=4$) & \textbf{32.51} \goldmedal & \textbf{0.202} \goldmedal & 0.400 \silvermedal & 604k \silvermedal & \textbf{13.49x} \goldmedal \\ 
    \cline{2-7}
    
    & Non-neural (ref.) & 
    32.01 & 
    0.225 & 
    0.432 &  
    391k & 
    28.32x \\

    \hline 
    \hline 
	\multirow{1}{5em}{cat \newline $|V| = 33$k }

    & NeuTex & 31.56 & 0.338 & 0.336 & 793k & 1.0x \\ 
    & TF+RFF & 34.33 & \textbf{0.162} \goldmedal & 0.247 \silvermedal & 331k \silvermedal & 1.96x \\ 
    & INF & \textbf{34.76} \goldmedal & 0.166 \silvermedal & \textbf{0.202} \goldmedal & 36430k & 3.07x \\\cline{2-7}

    & Ours ($\lambda_{reg}=0$) & 33.27 & 0.305 & 0.453 & 166k & 13.33x \\
    & Ours ($d\uparrow$) & 34.65 \silvermedal & 0.191 & 0.295 & 411k & 11.94x \silvermedal \\
    & Ours ($d=4$) & 34.23 & 0.238 & 0.349 & \textbf{166k} \goldmedal & \textbf{13.33x} \goldmedal \\ 
    
    \cline{2-7}
    & Non-neural (ref.) & 33.01  & 0.400 & 0.775 & 123k & 32.52x \\ 
    
    \hline
    \end{tabular}
    \vspace{0.3cm}

    \caption{Texture reconstruction from multi-view images. Our multi-resolution feature encoding significantly improves inference speed compared to state-of-the-art neural methods while maintaining similar reconstruction quality.
    Note that DSSIM and LPIPS are scaled by 100. 
    The trade-off for the speed-up is a slight increase in the number of parameters; however, we still have a significantly smaller model compared to INF, which generally produces the best results.
    While the non-neural reference baseline has the fastest evaluation time, it shows significantly decreased reconstruction quality, particularly for the coarser mesh.
    For our method, using the regularizer shows a substantial improvement in reconstruction quality.
    While for coarser meshes like the \textit{cat}, we obtain 
    better reconstruction results
    for an increase in feature dimension ($d\uparrow$) to $d=10$, we did observe slight overfitting for finer meshes like the \textit{human}.
    }
    \label{tab:texture_recon_quant}
    \end{table*}

\newcommand{\imgwidthtexrecon}{0.16}

\newcommand{\texreconspyboxes}{
\spy [\greencol] on (0.15,2.9) in node [left] at (-0.5,2.5); %
\spy [\redcol] on (0.5, 2.5) in node [left] at (-0.5,1); %
}
\newcommand{\magnifactor}{2}
\begin{figure}[ht]
    \centering
  \begin{subfigure}{\imgwidthtexrecon\textwidth}
        \begin{tikzpicture}[spy using outlines={magnification=\magnifactor, size=1cm, connect spies}, remember picture]
        \node[anchor=south] at (0,0) {\includegraphics[width=\textwidth]{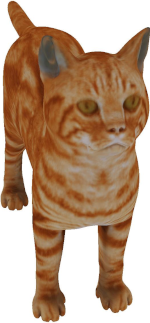}};
        \texreconspyboxes
        \end{tikzpicture}
    \caption{NeuTex}
    \label{fig:neutex_qual}
  \end{subfigure} \hfill
  \begin{subfigure}{\imgwidthtexrecon\textwidth}
        \begin{tikzpicture}[spy using outlines={magnification=\magnifactor, size=1cm, connect spies}, remember picture]
        \node[anchor=south] at (0,0) {\includegraphics[width=\textwidth]{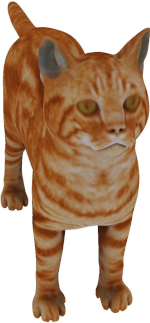}};
        \texreconspyboxes
        \end{tikzpicture}
    \caption{TF+RFF}
  \end{subfigure} \hfill
  \begin{subfigure}{\imgwidthtexrecon\textwidth}
        \begin{tikzpicture}[spy using outlines={magnification=\magnifactor, size=1cm, connect spies}, remember picture]
        \node[anchor=south] at (0,0) {\includegraphics[width=\textwidth]{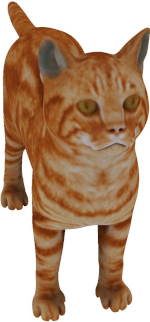}};
        \texreconspyboxes
        \end{tikzpicture}
    \caption{INF}
  \end{subfigure} \hfill
\begin{subfigure}{\imgwidthtexrecon\textwidth}
        \begin{tikzpicture}[spy using outlines={magnification=\magnifactor, size=1cm, connect spies}, remember picture]
        \node[anchor=south] at (0,0) {\includegraphics[width=\textwidth]{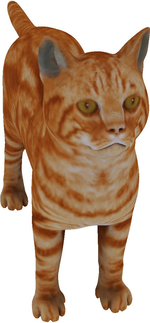}};
        \texreconspyboxes
        \end{tikzpicture}
    \caption{Ours}
\end{subfigure} \hfill
\begin{subfigure}{\imgwidthtexrecon\textwidth}
        \begin{tikzpicture}[spy using outlines={magnification=\magnifactor, size=1cm, connect spies}, remember picture]
        \node[anchor=south] at (0,0) {\includegraphics[width=\textwidth]{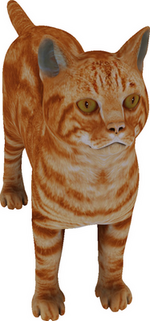}};
        \texreconspyboxes
        \end{tikzpicture}
    \caption{GT}
\end{subfigure}

    \caption{Qualitative results for texture reconstruction from multi-view images on the \textit{cat}. 
    Our method enables high-quality reconstructions, matching state-of-the-art methods in visual fidelity while offering a significant speedup. Because the baseline methods are based on frequency encodings, they lead to an over-smoothening of intricate details around the eye, which only our method can capture. Furthermore, NeuTex is unable to capture spatially fast changing color like on the mouth of the \textit{cat} and shows distortions inside the ear.
    }
    \label{fig:texture_recon_qualitative}
\end{figure}

We follow the experimental setup of Intrinsic Neural Fields \cite{Koestler22IntrinsicNeuralFields}, initially proposed by Oechsle et al. \cite{Oechsle19TextureFields} and use the same dataset.
The input to all methods is a set of five 512x512 pixel posed images alongside their respective intrinsic and extrinsic camera matrices. Additionally, the 
triangle mesh of the object is given.
In the preprocessing stage, we compute the ray-mesh intersection for each pixel 
to obtain the corresponding point on the mesh. We use the Euclidean embedding of this point as input for TF and use barycentric coordinates in the respective triangle for the other methods. 
Additionally, for INF, the values of the eigenfunctions of the Laplace-Beltrami operator (LBO) must be computed at the mesh vertices. We would like to point out that this computation can take \emph{hours}, depending on the mesh size, while the other methods do not require any preprocessing. For the training, we employ an L1 loss on the color values. We render 200 images for evaluation and report PSNR, DSSIM\footnote{For better readability we use the Structural Dissimilarity: $\text{DSSIM}= (1 - \text{SSIM})/2$ } \cite{WangBSS04SSIM} and LPIPS \cite{ZhangIESW18LPIPS}. All experiments were done on an Nvidia A10G Tensor Core with 24GiB of memory. 

The results in \cref{tab:texture_recon_quant} and \cref{fig:texture_recon_qualitative} show that our model achieves results on par with the state-of-the-art while showing a notable speed-up compared to previous approaches.

\paragraph{Reconstruction Quality}
\cref{tab:texture_recon_quant} shows that our method yields quantitative results that are on par with the other methods for the reconstruction quality.
As can be seen in \cref{fig:texture_recon_qualitative}, we even achieve a noticeable qualitative improvement for some regions, while other regions are slightly worse. We found that areas of high quality often correspond with a finer mesh in that area, which is to be expected since, for those regions, more features are available.

\paragraph{Inference Time}
To benchmark the inference speed, we measure the evaluation time of the encoding stage and the MLP for $2^{15}$ points for each model. Note that this does not include the ray mesh intersection or the computation of the eigenfunctions. To reduce variance, we average the results over 300 measurements of the same batch. \cref{tab:texture_recon_quant} shows a significant speed-up of our method compared to the other methods, which we attribute to the much smaller MLP size. Also, it is important to note that our encoding strategy is computationally lightweight since it involves a single matrix multiplication with low dimensions due to the small feature size. As expected, the non-neural baseline is the fastest.

\paragraph{Number of Parameters}
For the number of parameters, we count all parameters that are necessary to evaluate the model. This includes learnable components like the weights of the network and the features, as well as non-learnable ones like the Fourier feature matrix and the LBO eigenfunctions values. We observe that the trade-off for the evaluation speed of our method is a slightly increased number of parameters\footnote{As a reference: a 512x512 3-channel color image consists of over 786k pixel values.} due to the feature matrices. This observation is similar to feature-grid-based methods in the Euclidean space.

\subsection{Ablation Study}
\label{sec:ablation}

\subsubsection*{Multi-Resolution Features}

\begin{figure}
    \centering
    \input{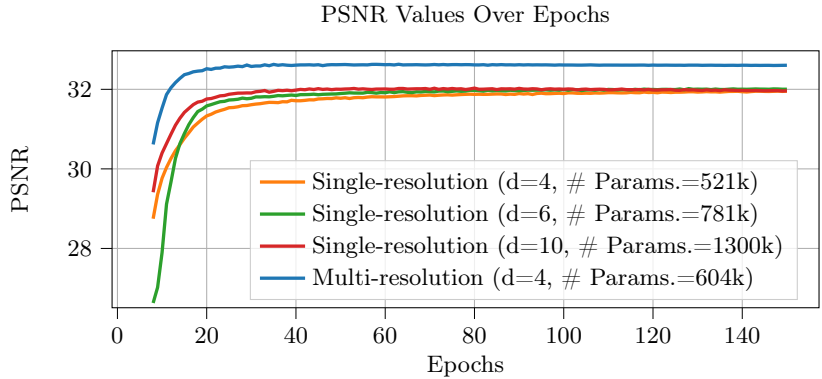}
    \caption{Validation PSNR over training time in epochs. Our multi-resolution feature encoding leads to higher reconstruction quality despite having fewer parameters than a single-resolution encoding. For the latter, we use the finest resolution and arrange parameters the same way as in the multi-resolution approach. 
    While a higher feature dimension $d$ leads to faster convergence in the single-resolution setting, it does not improve the reconstruction quality despite the increased number of parameters.
    }
    \label{fig:plot-multires}
\end{figure}

We accumulate information distributed over multiple resolutions of our mesh to enable different scales of locality. This way, coarse resolutions can capture more global information, and fine resolutions can act as correcting terms for local changes. 
\cref{fig:plot-multires} shows the effectivity of our multi-resolution approach and demonstrates that a single-resolution setup struggles to achieve high visual fidelity, even if the number of parameters exceeds the multi-resolution approach. It confirms that our multi-resolution approach allows an effective sharing of global information on different scales and enables reconstructions of higher quality while maintaining rapid convergence.

\subsubsection*{Regularization}
\newcommand{\spyboxes}{
\spy [\redcol] on (2.6,2.4) in node [left] at (4.5,1);
\spy [\bluecol] on (2.4,4.45) in node [left] at (5,3);
\spy [\greencol] on (2.6,0.5) in node [left] at (2,2);
}

\begin{figure}[ht!]
    \centering

    \begin{subfigure}{0.4\textwidth}
        \begin{tikzpicture}[spy using outlines={magnification=3, size=1cm, connect spies}, remember picture]
            \node[anchor=south west] at (0,0) {\includegraphics[width=\textwidth]{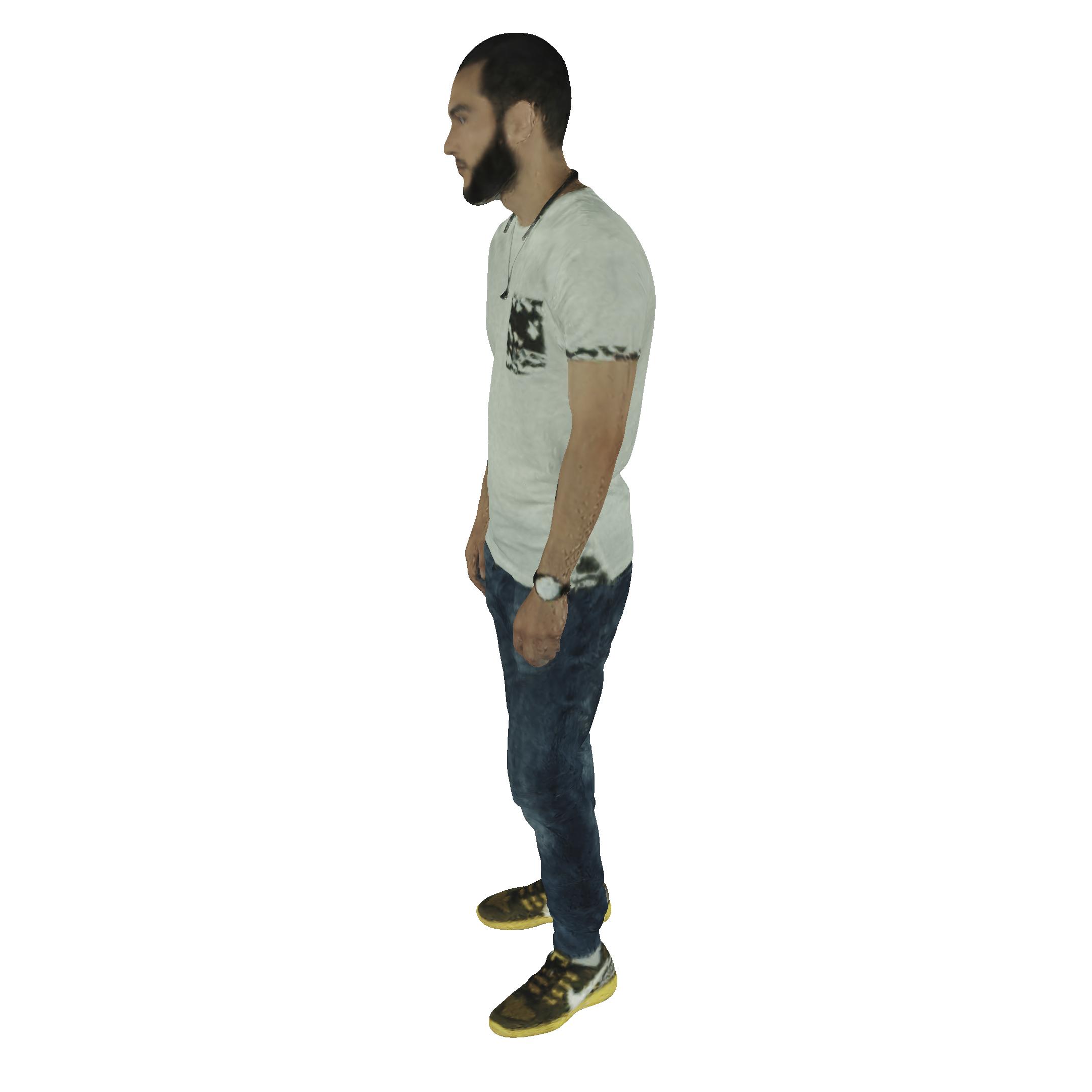}};
             \spyboxes
        \end{tikzpicture}
        \caption{Training without regularization}
    \end{subfigure}
    \begin{subfigure}{0.4\textwidth}
        \begin{tikzpicture}[spy using outlines={magnification=3, size=1cm, connect spies}, remember picture]
            \node[anchor=south west] at (0,0) {\includegraphics[width=\textwidth]{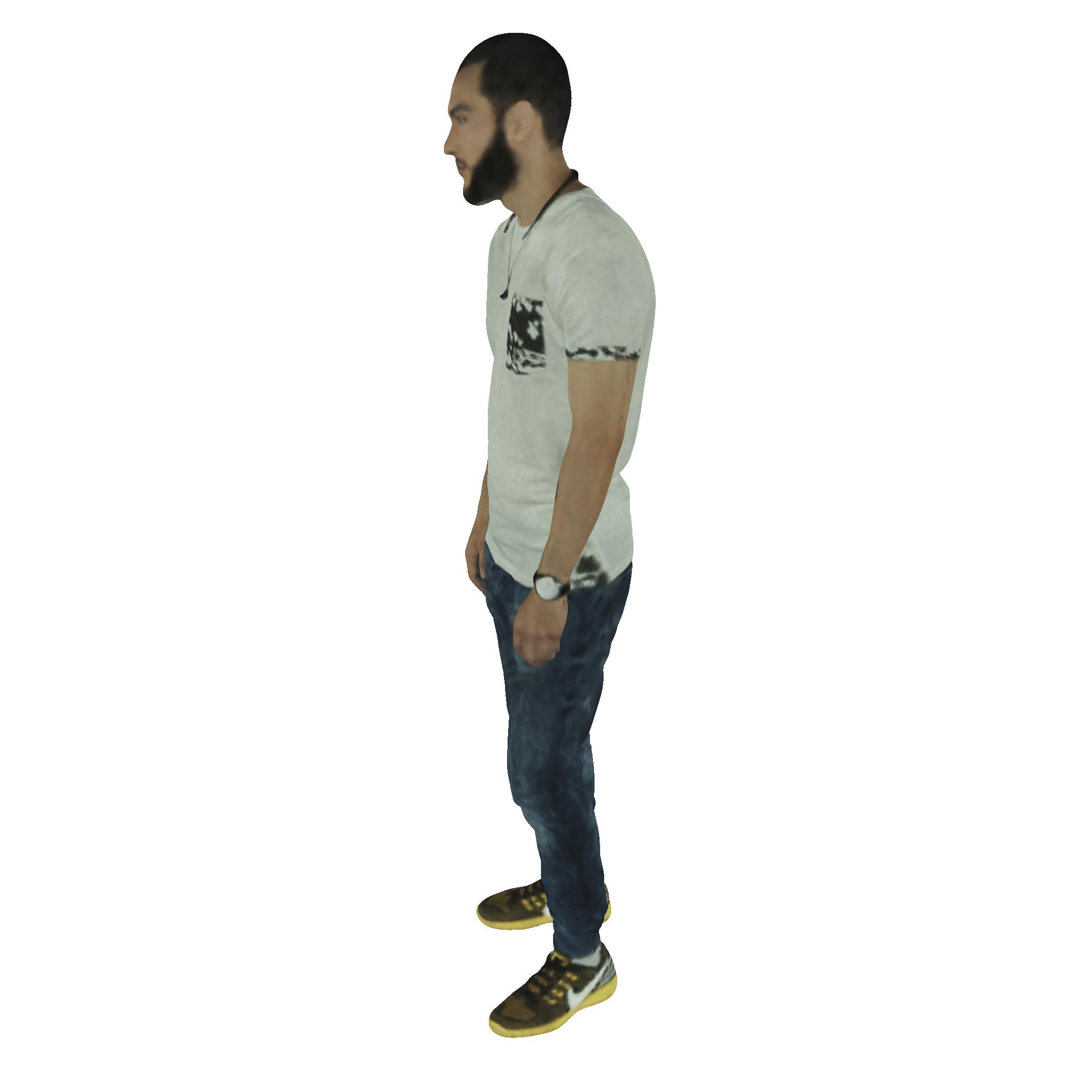}};
             \spyboxes
        \end{tikzpicture}
    \caption{Training with regularization}
    \end{subfigure}

    \caption{Qualitative results of our method for texture reconstruction with 
    and without the regularization based on the mesh Laplacian. 
    The results on the left, without the regularization, show visual artifacts around the ear, and on the shoe and arm. This is a direct result of sparse training data resulting in untrained feature vectors. Our regularization acts as a smoothing term that enables feature information to be diffused to the unsupervised areas, significantly reducing the noise.
    }
    \label{fig:regularization}
\end{figure}

Frequency encodings require the MLP to map encoded spatial information to function values. Due to strong coupling of spatial information in a single network, frequency encodings can interpolate for regions, where training data is sparse.
For our parametric encoding approach, working with fine meshes and sparse training data can lead to unsupervised latent codes, creating visual artifacts in our reconstructions. Our latent code regularization (\cref{eq:reg_laplacian}) uses the mesh Laplacian to circumvent this issue. \cref{fig:regularization} shows how our regularization reduces these artifacts significantly. The positive effect or our regularization scheme is also underlined quantitatively in \cref{tab:texture_recon_quant}.

\subsection{Deforming Meshes}
\begin{figure}[ht!]
  \begin{subfigure}{\imgwidthtexrecon\textwidth}
  
        \begin{tikzpicture}[remember picture]
        \node[anchor=south] (face) at (-2.5,0) {\includegraphics[width=\textwidth]{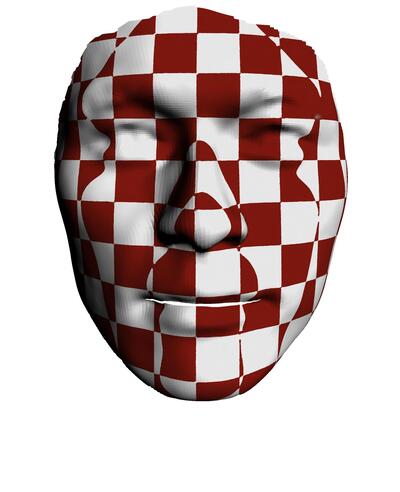}};
        \node[anchor=south] (elephant) at (-2.5, 2.5) {\includegraphics[height=1\textwidth]{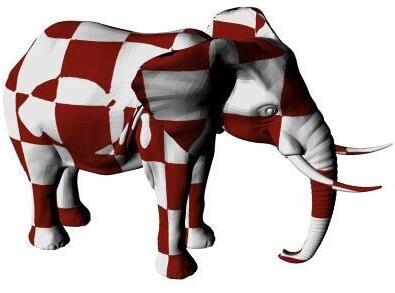}};

        \node (a) at (face.south) {Reference mesh};
        \draw[blue,thick,dotted] ($(elephant.north west)+(-0.1,0.6)$)  rectangle ($(face.south east)+(0.5,-0.6)$);

        \node[anchor=south] (face1) at (0.6,0) {\includegraphics[width=\textwidth]{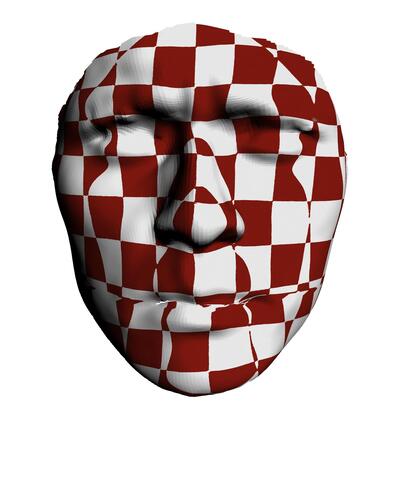}};
        \node[anchor=south] (elephant1) at (1.3, 2.5) {\includegraphics[height=1\textwidth]{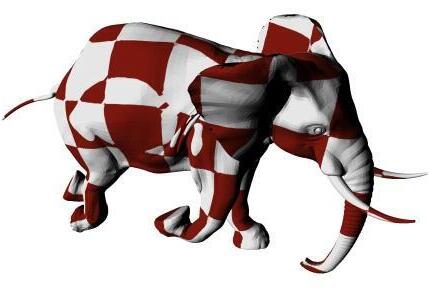}};

        \node[anchor=south] (face2) at (2.6,0) {\includegraphics[width=\textwidth]{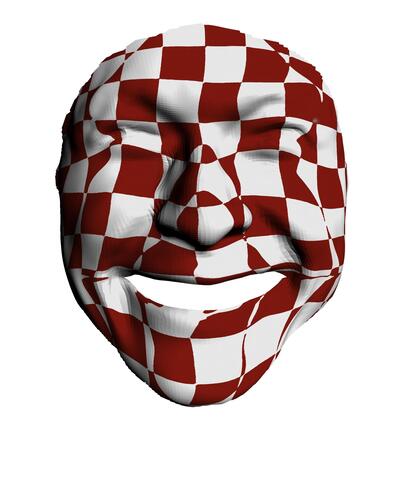}};        
        \node[anchor=south] (elephant2) at (4, 2.5) {\includegraphics[height=1\textwidth]{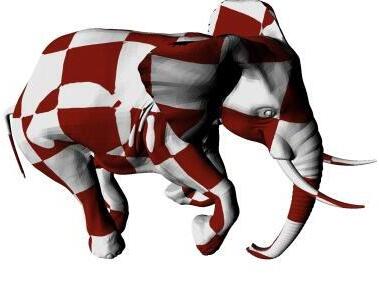}};     
        
        \node[anchor=south] (face3) at (4.6,0) {\includegraphics[width=\textwidth]{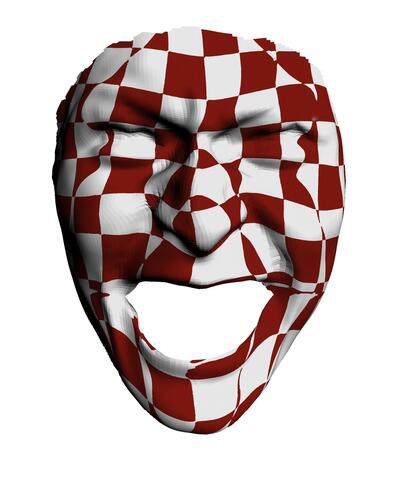}};        
        \node[anchor=south] (elephant3) at (6.7,2.5) {\includegraphics[height=1\textwidth]{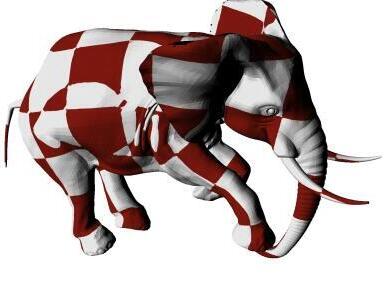}};   
        
        \node[anchor=south] (face4) at (6.6,0) {\includegraphics[width=\textwidth]{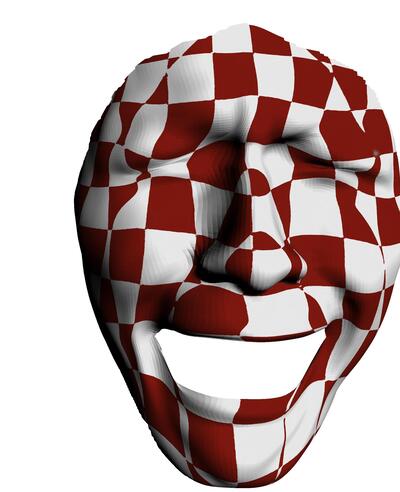}};        

        \node (b) at ($(face2.south)!0.5!(face3.south)$) {Rendering on deformed meshes};

        \draw[blue,thick,dotted] ($(elephant1.north west)+(-0.1,0.6)$)  rectangle ($(face4.south east)+(0.5,-0.6)$);
        
        \end{tikzpicture}
  \end{subfigure}

    \caption{Qualitative evaluation of texture represented by our method under mesh deformation. We train our network for a checker texture on the reference mesh (\textit{left}) and render images for various deformations.
    Since the mesh topology remains unchanged under deformations, and our features are intrinsic to the mesh, our representation is unaffected by the deformation and can be evaluated similarly to the reference configuration.
    The results show that the representation is consistent under deformations and produces no visible artifacts.
    }
    \label{fig:deformations_qualitative}
\end{figure}
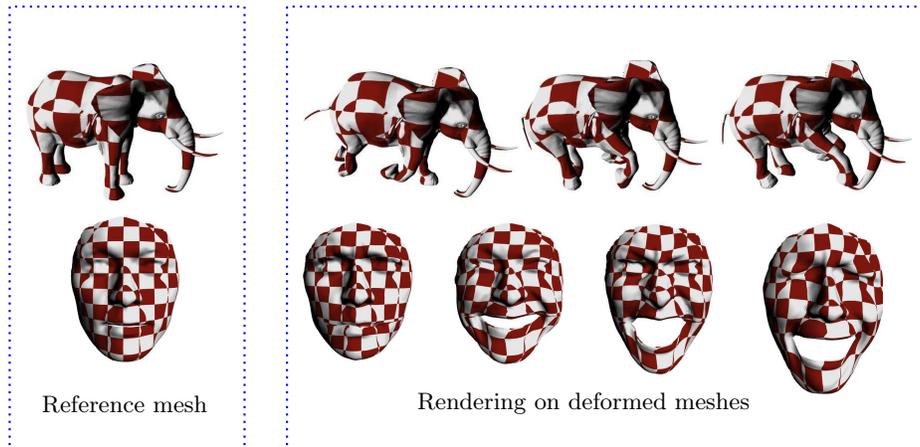

For many real-world applications of meshes that include animations, it is crucial that the representation of \eg texture can also be used under deformations of the mesh. Since our features are stored at the mesh vertices and are therefore intrinsic to the mesh, they are unaffected by deformations. Consequently, our method supports mesh deformations natively without any additional computational overhead.
While the same is true for other intrinsic encodings \cite{Koestler22IntrinsicNeuralFields}, extrinsic methods like \cite{Oechsle19TextureFields, tancik2020fourier} need to map the intersection point in the deformed configuration back to the reference configuration, which adds a slight computational overhead.
In \cref{fig:deformations_qualitative}, we show qualitatively that a texture represented by our method is consistent when the mesh is deformed and exhibits no noticeable artifacts. For the experiments, we use a checker texture on two meshes from the data provided by Sumner and Popovic \cite{sumner2004deformation}, which we learn directly on the respective reference mesh and evaluate on some of its deformations.

\subsection{BRDF Reconstruction from Images with Calibrated Lighting}
As a second application of our method, we consider the estimation of a spatially varying \emph{bidirectional reflectance distribution function} (BRDF) on a mesh. The BRDF $f(x, l, v)$ describes how much of the irradiance incident from the light direction $l\in\mathbb{S}^2$ is reflected in the viewing direction $v\in\mathbb{S}^2$. To obtain the total outgoing radiance $L_o(x, v)$ at position $x$ in the viewing direction $v$, the incoming irradiance $L_i(x, l)$ needs to be integrated over the hemisphere $\mathbb{H}$, which is known as the rendering equation
\begin{equation}\label{eq:rendering_eq}
    L_o(x, v) = \int_{\mathbb{H}} f(x, l, v) L_i(x, l) \cos\theta_l \,\mathrm{d}l.
\end{equation}

For the estimation of the BRDF, we consider the special case of a single, directional light. As a result, the irradiance $L_i$ is independent of the position $x$, and the rendering equation \cref{eq:rendering_eq} reduces to a single evaluation for the direction $l$ of the directional light. The simplified equation reads
\begin{equation}
    L_o(x, v) = f(x, l, v) L_i \mathbb{I}_s(x, l) \cos\theta_l,
\end{equation}
where we have introduced the indicator function $\mathbb{I}_s(x, l)$ to account for cast shadows.

Several models have been proposed to parametrize the BRDF. In this work, we choose the isotropic variant of the state-of-the-art Disney BRDF, which uses 12 parameters with values in the unit interval \cite{burley2012physically}. We compare against texture fields and intrinsic neural fields and modify the methods to predict these parameters spatially varying on the mesh by adjusting the output dimension accordingly. For Texture Fields, we found that an exponential learning rate scheduler with $\gamma = 0.9$ is necessary to prevent overfitting. All other methods remain unchanged. We employ a batch size of $b=2^{14}$ color values under a given light direction for the training of all methods.

\begin{figure}[ht!]
    \centering
    \newcommand\imgwidthbrdf{0.24}
  \begin{subfigure}{\imgwidthbrdf\textwidth}
    \includegraphics[width=\linewidth]{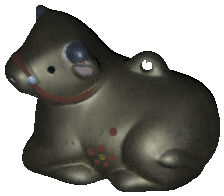}
    \caption{TF+RFF}
  \end{subfigure}
  \begin{subfigure}{\imgwidthbrdf\textwidth}
  \includegraphics[width=\linewidth]{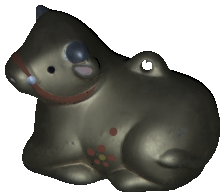}
    \caption{INF}
  \end{subfigure}
  \begin{subfigure}{\imgwidthbrdf\textwidth}
  \includegraphics[width=\linewidth]{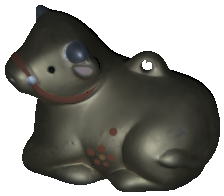}
    \caption{Ours}
  \end{subfigure}
\begin{subfigure}{\imgwidthbrdf\textwidth}
\includegraphics[width=\linewidth]{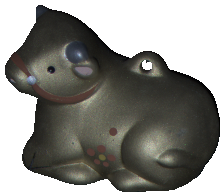}
    \caption{GT}
\end{subfigure}

    \caption{Qualitative Results of the BRDF reconstruction for the \emph{cow} of the DiLiGenT-MV real-world dataset. The renderings show, that our method is able to reconstruct sharp boundaries in the material and yields results that are practically indistinguishable from the baseline methods, while being almost an order of magnitude faster.
    }
    \label{fig:brdf_recon_qualitative}
\end{figure}
\begin{table*}[ht!]  %
  \centering  %
  
    \begin{tabular}{ | l | l l l l l |} 
  \hline
    Method & PSNR $\uparrow$ & DSSIM $\downarrow$ & LPIPS $\downarrow$ & \# Params. 
    $\downarrow$ & Speedup $\uparrow$  \\	\hline 

TF+RFF & 42.13 & 0.6718 & \textbf{1.50} \goldmedal & \textbf{332k} \goldmedal & 1.00x  \\ 	%
INF & \textbf{42.21} \goldmedal & \textbf{0.666} \goldmedal & 1.53 \silvermedal & 204930k & 1.08x \silvermedal  \\ 	%
Ours & 42.17 \silvermedal & 0.6700 \silvermedal & 1.60 & 930k \silvermedal & \textbf{7.58x} \goldmedal  \\ 	%

    \hline
    \end{tabular}
    \vspace{0.3cm}

\caption{Results of the BRDF reconstruction. All metrics are averaged over the experiments for the 5 objects of the  DiLiGenT-MV real-world dataset. Note that DSSIM and LPIPS are scaled by 100. Our method archives again similar reconstruction quality to the other methods while maintaining the large speed-up observed previously.}
\label{tab:brdf_recon_quant}
\end{table*}

We use the established DiLiGenT-MV real-world dataset for our experiments \cite{LiZ20DiligentDataset}. It contains HDR images of 5 objects with complex reflective behavior. The images are taken from 20 calibrated views, captured under 96 calibrated directional lights each. Moreover, the dataset contains meshes of the objects.
For training, we use 10 views with 30 lights each and evaluate the models on 20 lights for each of the remaining 10 views. We follow \cite{Mildenhall22NerfInTheDark} and combine an L1 loss with a gamma correction function applied to the color values. Since the dataset contains HDR images, very bright regions would otherwise dominate the loss, and information from darker regions would be suppressed. We refer to the supplement for more details.
The results in \cref{tab:brdf_recon_quant} and \cref{fig:brdf_recon_qualitative} show, that for BRDF estimation, we achieve similar reconstruction quality to the other methods while maintaining the significant speed-up in evaluation time. For more qualitative results, we refer to the supplement.

\section{Conclusion}
\label{sec:conclusion}
We have introduced MeshFeat as a multi-resolution parametric feature encoding for neural fields on meshes. The approach adapts the idea of parametric feature grids from the Euclidean space to meshes. Compared to existing approaches, this enables us to work with considerably smaller MLPs, leading to a significant speed-up. Our proposed multi-resolution approach is based on mesh simplification and enables the effective sharing of global feature information. This allows for higher reconstruction quality than a single-resolution approach while simultaneously requiring a smaller feature dimension. We have demonstrated the computational efficiency of our encoding for texture and BRDF representation on meshes, where we achieve reconstruction results that are on par with competing methods based on frequency-based encodings with significantly longer evaluation times. Additionally, we showcased our method's native applicability to signals on deforming meshes. 
We identify a more texture-adaptive multi-resolution feature approach as a promising direction for future research. 
We hope our work provides valuable insights applicable to various areas, including virtual and augmented reality, animations, and computer graphics engines.

\paragraph{Acknowledgments.} We thank Christian Koke for helpful discussions. This work was supported by ERC Advanced Grant SIMULACRON.

\bibliographystyle{splncs04}
\bibliography{biblio}

\newpage
\appendix
\section*{Appendix}
In this supplementary material, we give additional details on the multi-resolution strategy in \cref{sec:supp_add_multires}, on the training in \cref{sec:supp_add_training}, as well as additional experimental results in \cref{sec:supp_add_exps}. The latter includes additional renderings for all objects as well as quantitative results for the individual objects of the BRDF reconstruction. Moreover, we present a qualitative analysis of the influence of the different resolutions of our multi-resolution approach in \cref{sec:supp_vis_multi-res} and further analysis of the hyperparameters in \cref{sec:supp_add_analysis_hyperparams}.
\section{Mesh Simplification-Based Multi-Resolution Strategy}
\label{sec:supp_add_multires}
\def\xOffset{3.5cm}

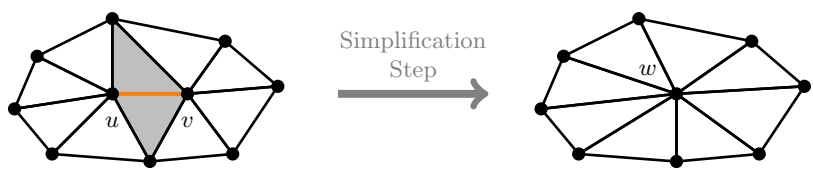
\begin{figure}
    \centering
\tikzset{
    vertex/.style={
        circle,fill=black,inner sep=0pt,minimum size=5pt
    },
    shiftLeft/.style={
        transform canvas={xshift=-\xOffset}
    },
    shiftRight/.style={
        transform canvas={xshift=\xOffset}
    }
}
\begin{tikzpicture}[
    label distance=0.8mm,
    line width=1pt
]

\coordinate (U) at (-0.5,0) {};
\coordinate (V) at (0.5,0) {};
\coordinate (W) at (0,0) {};

\coordinate (A) at (-0.5,1) {};
\coordinate (B) at (0,-0.9) {};
\coordinate (C) at (-1.5,0.5) {};
\coordinate (D) at (-1.8,-0.2) {};
\coordinate (E) at (-1.3,-0.8) {};
\coordinate (F) at (1.1,-0.8) {};
\coordinate (G) at (1.7,0.1) {};
\coordinate (H) at (1.0,0.7) {};

\draw[shiftLeft, fill=lightgray] (U) -- (V) -- (A) -- (U);
\draw[shiftLeft, fill=lightgray] (U) -- (V) -- (B) -- (U);
\draw[shiftLeft] (U) -- (A) -- (C) -- (U);
\draw[shiftLeft] (U) -- (C) -- (D) -- (U);
\draw[shiftLeft] (U) -- (D) -- (E) -- (U);
\draw[shiftLeft] (U) -- (E) -- (B) -- (U);
\draw[shiftLeft] (V) -- (B) -- (F) -- (V);
\draw[shiftLeft] (V) -- (F) -- (G) -- (V);
\draw[shiftLeft] (V) -- (G) -- (H) -- (V);
\draw[shiftLeft] (V) -- (H) -- (A) -- (V);
\draw[shiftLeft, color=orange, line width=1.5pt] (U) -- (V);

\draw[shiftLeft] (U)node[vertex,label=below:{$u$}]{};
\draw[shiftLeft] (V)node[vertex,label=below:{$v$}]{};
\draw[shiftLeft] (A)node[vertex]{};
\draw[shiftLeft] (B)node[vertex]{};
\draw[shiftLeft] (C)node[vertex]{};
\draw[shiftLeft] (D)node[vertex]{};
\draw[shiftLeft] (E)node[vertex]{};
\draw[shiftLeft] (F)node[vertex]{};
\draw[shiftLeft] (G)node[vertex]{};
\draw[shiftLeft] (H)node[vertex]{};

\draw[shiftRight] (W) node[vertex,label=north west:{$w$}]{};
\draw[shiftRight] (A)node[vertex]{};
\draw[shiftRight] (B)node[vertex]{};
\draw[shiftRight] (C)node[vertex]{};
\draw[shiftRight] (D)node[vertex]{};
\draw[shiftRight] (E)node[vertex]{};
\draw[shiftRight] (F)node[vertex]{};
\draw[shiftRight] (G)node[vertex]{};
\draw[shiftRight] (H)node[vertex]{};

\draw[shiftRight] (W) -- (A) -- (C) -- (W);
\draw[shiftRight] (W) -- (C) -- (D) -- (W);
\draw[shiftRight] (W) -- (D) -- (E) -- (W);
\draw[shiftRight] (W) -- (E) -- (B) -- (W);
\draw[shiftRight] (W) -- (B) -- (F) -- (W);
\draw[shiftRight] (W) -- (F) -- (G) -- (W);
\draw[shiftRight] (W) -- (G) -- (H) -- (W);
\draw[shiftRight] (W) -- (H) -- (A) -- (W);

\draw[->, line width=3pt, color=gray] (-1, 0) -- node[above, align=center]{Simplification \\ Step} (1, 0);

\end{tikzpicture}
\caption{
Visualization of a single simplification step. The orange edge has been chosen by the algorithm to be contracted. The two adjacent vertices $u$ and $v$ are collapsed into the vertex $w$. The two gray triangles are removed in this process.
Figure adapted from \cite{GarlandH97SurfaceSimplification}.
}
\label{fig:mesh_simplification}
\end{figure}

As described in the main text, we use the mesh simplification algorithm by Garland and Heckbert \cite{GarlandH97SurfaceSimplification} to construct our multi-resolution approach. The algorithm iteratively contracts vertex pairs until a specified target number of vertices is reached. The pairs to be contracted are selected based on a geometric error, as shown in the original work. 
See \cref{fig:mesh_simplification} for a visualization of a single contraction step on an edge.

By using this simplification algorithm, we obtain the sequence of meshes $((V^{(i)}, F^{(i)}))_i$ with the specified target resolutions.
Recall that the multi-resolution strategy is then based on the mapping 
\begin{equation}
    m^{(i)}: V\rightarrow V^{(i)}\quad v\mapsto m^{(i)}(v)=v^{(i)}
    \label{eq:supp_multi_res_mapping}
\end{equation}
(\iftoggle{arxiv}{\cref{eq:multi-res-mapping}}{Eq. (1)}
in the main text)
that assigns a vertex $v\in V$ in the original to mesh the vertex $v^{(i)}\in V^{(i)}$ in resolution $i$ to which $v$ was collapsed in the decimation algorithm. Consider \cref{fig:mesh_simplification} as an example and assume that the mesh on the left is the original resolution and the mesh on the right is the $i$-th resolution. In that case, $m^{(i)}$ would map both vertices $u$ and $v$ to $w$, \ie $m^{(i)}(u)= w$ and $m^{(i)}(v)=w$. The idea works analogously for multiple simplification steps between the resolutions.

We interpolate the features \emph{in the original resolution}. To do so, we aggregate the features of the different resolutions based on the mapping in \cref{eq:supp_multi_res_mapping}: For each vertex in the original resolution, we query to which vertex in the coarser resolutions it was collapsed and retrieve the respective features. To obtain the final multi-resolution feature for this vertex, we sum the features from all resolutions according to 
\iftoggle{arxiv}{\cref{eq:feature_acc}}{Eq. (2)}
in the main text.
Note that we do \emph{not} compute a geometric mapping between the resolutions but only use the connectivity information yielded by the mesh simplification algorithm and contained in the mapping. The interpolation of the features is performed based on the Barycentric coordinates of the respective triangle in the original mesh, see
\iftoggle{arxiv}{\cref{eq:bary_interp}}{Eq. (3)}
in the main text.

\section{Additional Training Details}
\label{sec:supp_add_training}

\subsection{Training Details for Texture Reconstruction}
\begin{figure}[ht]
    \centering

    \begin{subfigure}{0.19\textwidth}
        \includegraphics[width=\linewidth]{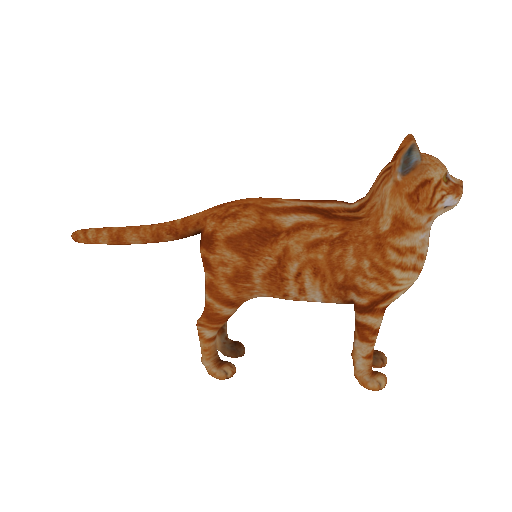}
    \end{subfigure}
    \hfill
    \begin{subfigure}{0.19\textwidth}
        \includegraphics[width=\linewidth]{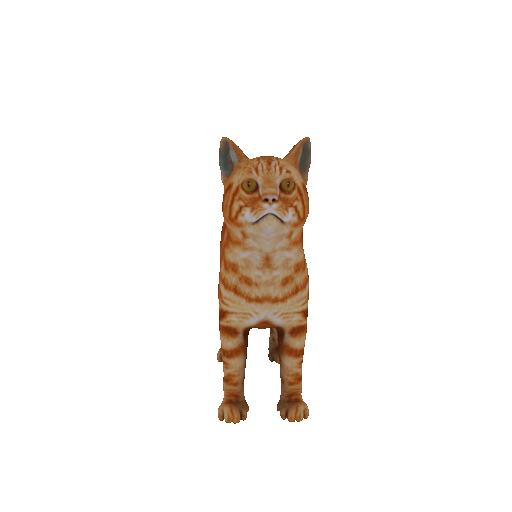}
    \end{subfigure}
    \hfill
    \begin{subfigure}{0.19\textwidth}
        \includegraphics[width=\linewidth]{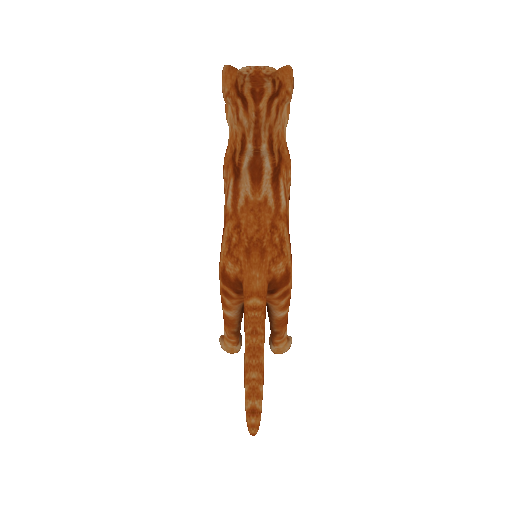}
    \end{subfigure}
    \hfill
    \begin{subfigure}{0.19\textwidth}
        \includegraphics[width=\linewidth]{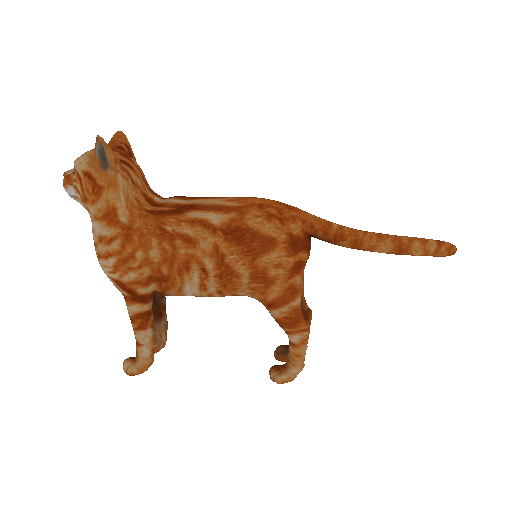}
    \end{subfigure}
    \hfill
    \begin{subfigure}{0.19\textwidth}
        \includegraphics[width=\linewidth]{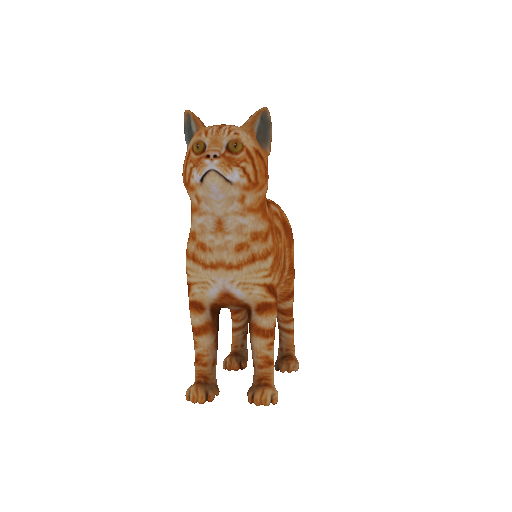}
    \end{subfigure}

    \vspace{10pt} %

    \begin{subfigure}{0.19\textwidth}
        \includegraphics[width=\linewidth]{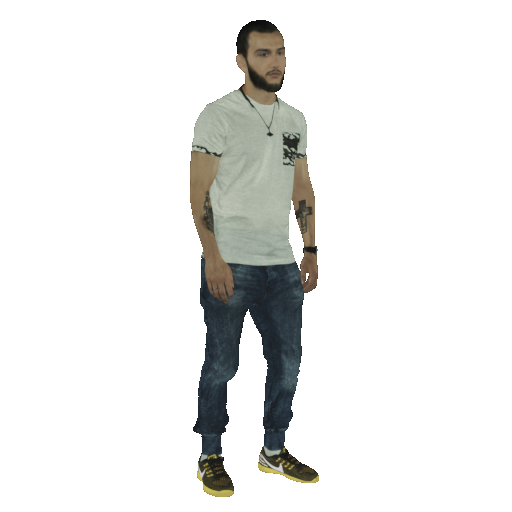}
    \end{subfigure}
    \hfill
    \begin{subfigure}{0.19\textwidth}
        \includegraphics[width=\linewidth]{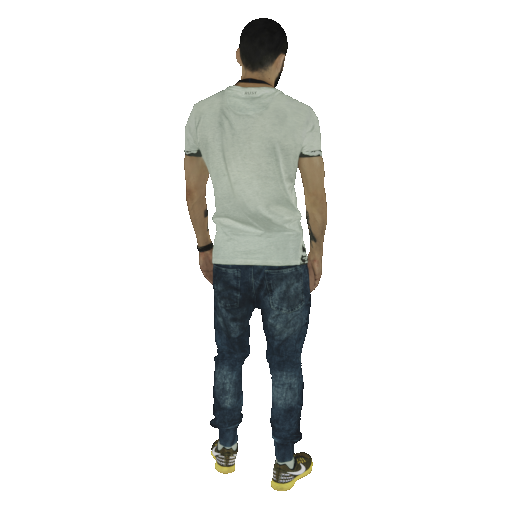}
    \end{subfigure}
    \hfill
    \begin{subfigure}{0.19\textwidth}
        \includegraphics[width=\linewidth]{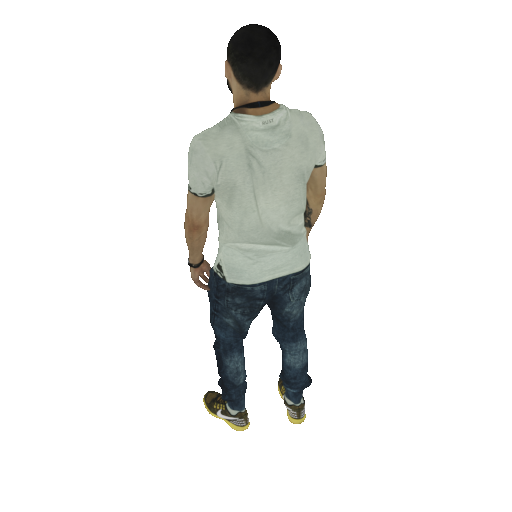}
    \end{subfigure}
    \hfill
    \begin{subfigure}{0.19\textwidth}
        \includegraphics[width=\linewidth]{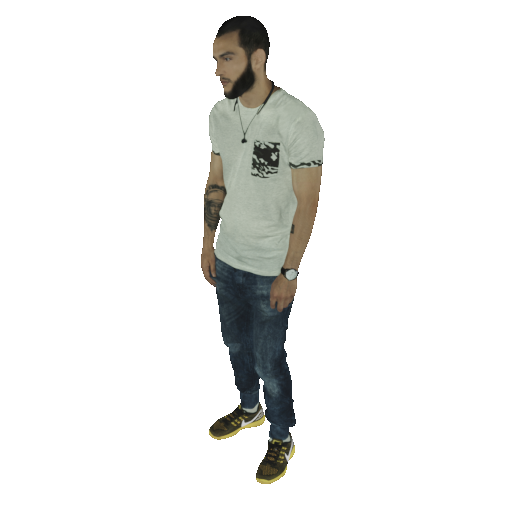}
    \end{subfigure}
    \hfill
    \begin{subfigure}{0.19\textwidth}
        \includegraphics[width=\linewidth]{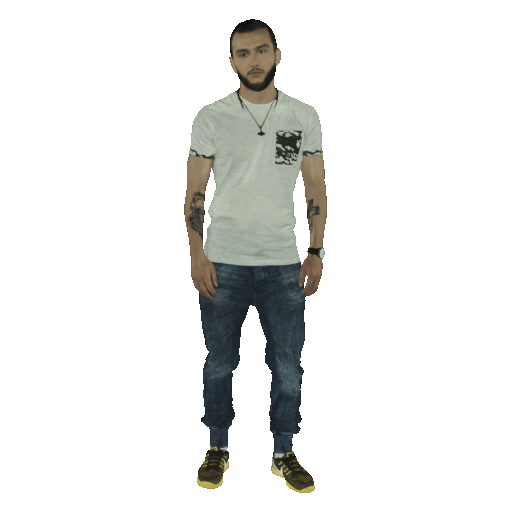}
    \end{subfigure}

    \caption{Training views for the cat and human dataset used for the texture reconstruction experiments. Note that we use the same views as done in the training by \cite{Koestler22IntrinsicNeuralFields}.}
    \label{fig:training_views}
\end{figure}

For the texture reconstruction experiments, we employ the same dataset as in \cite{Oechsle19TextureFields, Koestler22IntrinsicNeuralFields}, which includes the mesh and multiple views of the cat and human object. We use the same views as done in \cite{Koestler22IntrinsicNeuralFields} to make the comparison as direct as possible. These contain a set of 5 training, 100 validation and 200 test 512x512 views. The training images can be seen in \cref{fig:training_views}. 

\subsection{DiLiGenT-MV Dataset}
For the experiments on the BRDF reconstruction, we use the DiLiGenT-MV real-world dataset, which contains HDR images of 5 objects, taken from calibrated viewpoints under calibrated lighting conditions \cite{LiZ20DiligentDataset}. The triangle meshes provided with the dataset contain an excessively large number of vertices, leading to a significantly prolonged computation time for the LBO eigenfunctions for INF and an increased occurrence of unsupervised vertices in our method. Therefore, we reduce the number of vertices from roughly $10^6$ to about $2\cdot10^5$. To stay consistent with the simplified mesh, we compute the normals of the simplified mesh for the rendering rather than using the normal maps included in the dataset.

\subsection{Loss for the BRDF Estimation}
To avoid a dominant influence of the bright regions on the loss, we use a gamma mapping to transform the RGB values from linear to sRGB space before applying the L1 loss, as proposed by \cite{Mildenhall22NerfInTheDark}. Hence, the loss formulation reads
\begin{equation}
    \mathcal{L}_\text{data} = \frac{1}{N}\sum_{i=1}^N |\gamma(I_o(x, v)) - \gamma(I_{GT}(x, v))|,
\end{equation}
where $I_o(x, v)$ is the rendered color and $I_{GT}(x, v)$ is the ground truth color of the pixel corresponding to $x$ and $v$ in linear color space.
We use the following standard formula for the gamma mapping $g:[0,1]\rightarrow[0,1],~c_{\mathrm{lin}}\mapsto c_{\mathrm{sRGB}}$ described in \cite{akenine2019realTimeRendering}:
\begin{equation}
    g(c_{\mathrm{lin}}) = \begin{cases}
        \frac{323}{25}~c_{\mathrm{lin}} & \mathrm{if}~c_{\mathrm{lin}}\leq0.0031308 \\
        \frac{211}{200}~c_{\mathrm{lin}}^\frac{5}{12} - \frac{11}{200} & \mathrm{else } \\   
    \end{cases}
\end{equation}

\section{Additional Experimental Results}
\label{sec:supp_add_exps}

\subsection{Visualization of the Muli-Resolution Features}
\label{sec:supp_vis_multi-res}

Our multi-resolution strategy enables an effective sharing of common features. Coarser resolutions can learn global features, while finer resolutions act as a correcting term for details. We visualize this, by rendering the trained model with different resolution stages deactivated. More precisely, we modify the feature gathering described in 
\iftoggle{arxiv}{\cref{eq:feature_acc}}{Eq. (2)} 
in the main text, such that we do not sum the contributions over all resolutions but only over a subset of the resolutions. We start from the coarsest one and successively add finer resolutions. 

\begin{figure}[ht]
    \centering

    \scriptsize
  
    \newcommand{\mywidthx}{0.22\textwidth}  %
    \newcommand{\myheightx}{0.3\textwidth}  %
  
    \newcolumntype{X}{ >{\centering\arraybackslash} m{\mywidthx} } %

    \def\arraystretch{1} %
  \begin{tabular}{XXXX}
    \includegraphics[width=\mywidthx]{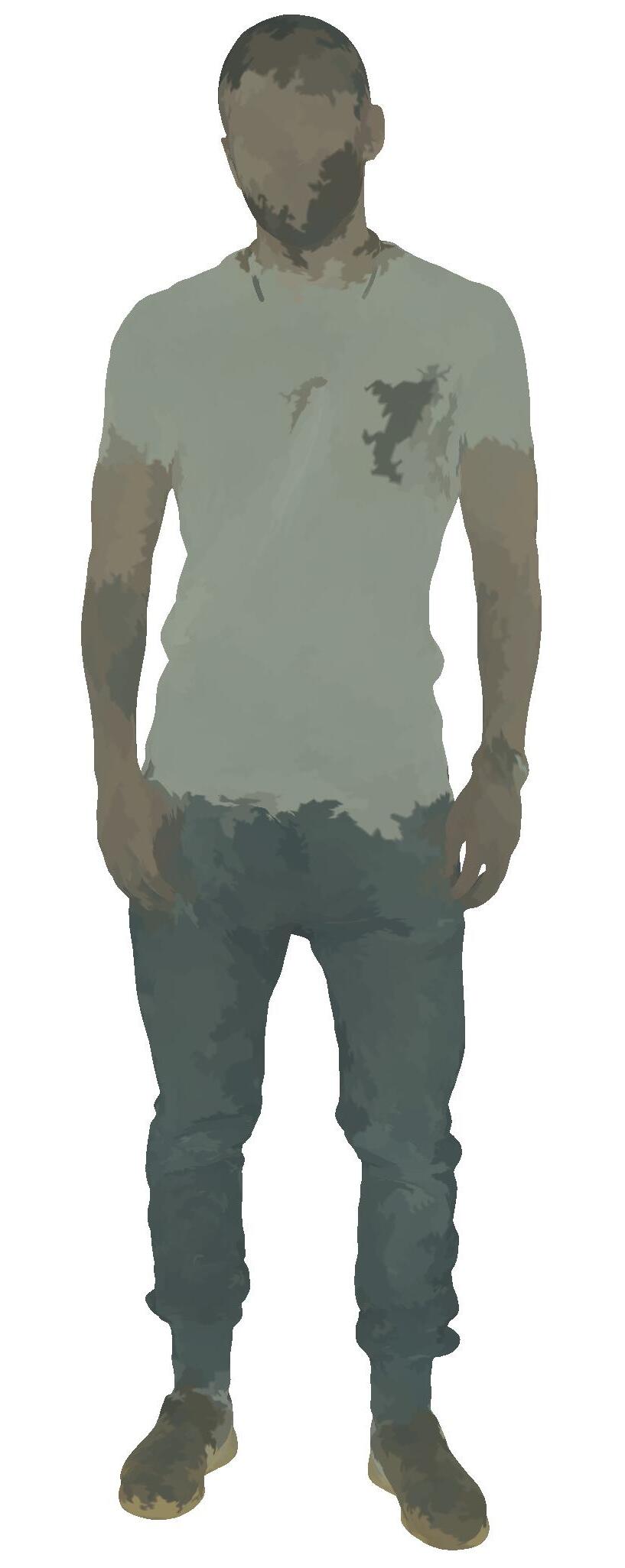} &
    \includegraphics[width=\mywidthx]{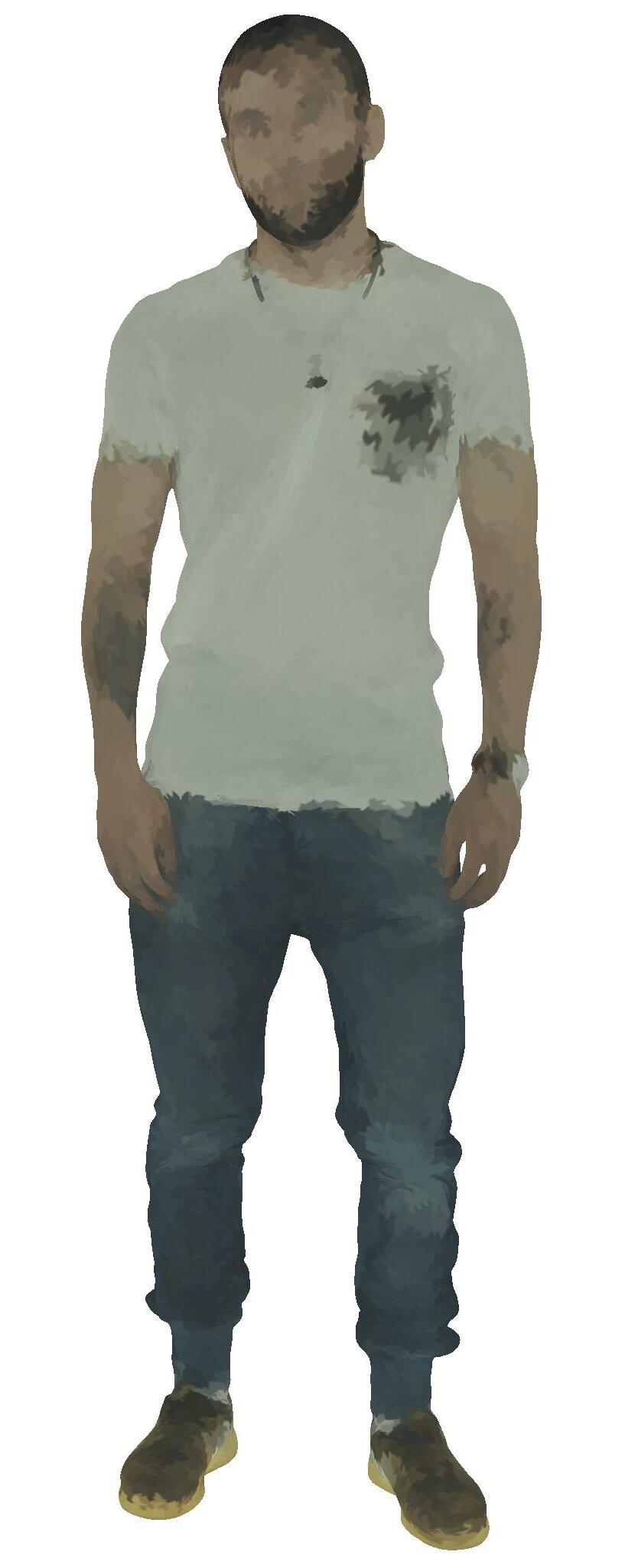} &
    \includegraphics[width=\mywidthx]{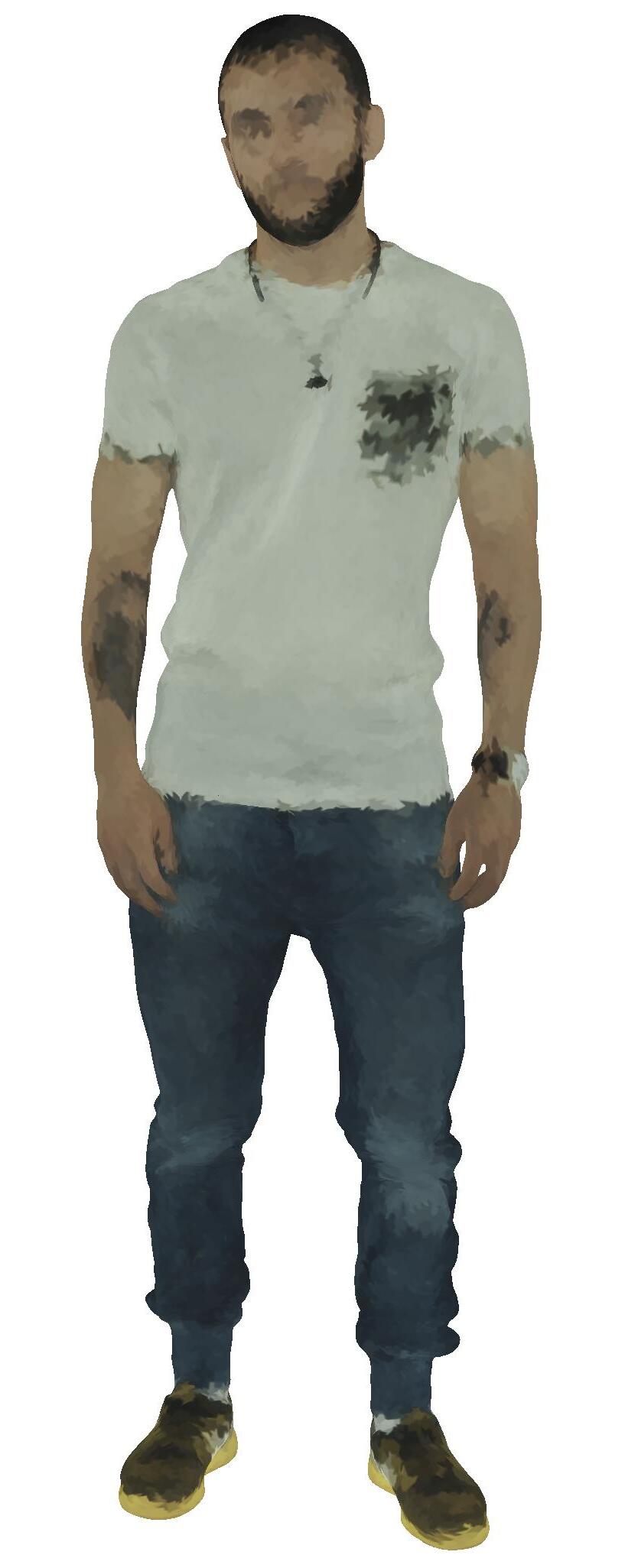} &
    \includegraphics[width=\mywidthx]{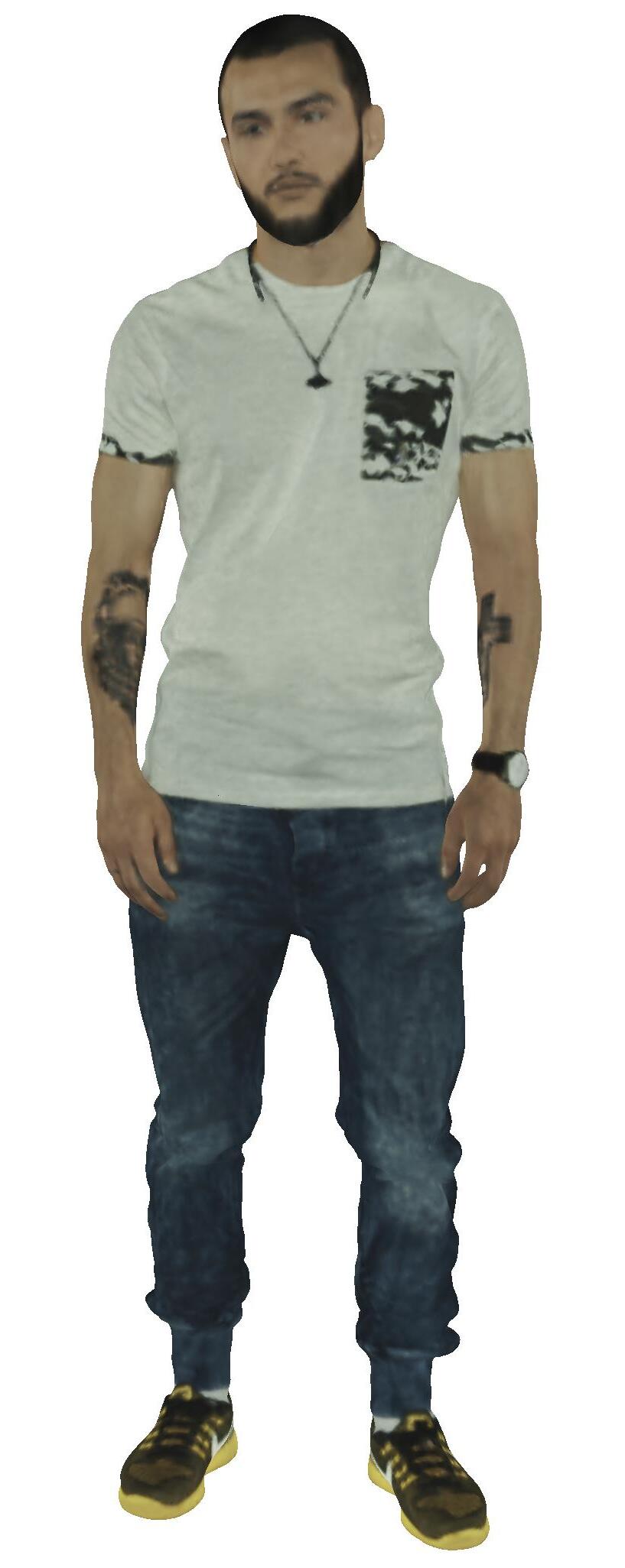} \\

    1 Resolution &
    2 Resolutions &
    3 Resolutions &
    All 4 Resolutions \\

  \end{tabular}

    \caption{
    Renderings with progressive unlocking of finer resolutions. The model was trained as detailed in the main text with 4 resolution levels $r^{(i)} \in \{1, 0.1, 0.05, 0.01\}$. From left to right we successively include finer resolutions in our feature gathering (\iftoggle{arxiv}{\cref{eq:feature_acc}}{Eq. (2)} in the main text), starting from only the coarsest resolution. The results show that our multi-resolution approach works as expected, and coarser resolutions (\textit{left}) capture global features, while finer resolutions (\textit{right}) enhance details in our representation. 
    }
    \label{fig:supp_multires_rendering}
\end{figure}

Qualitative results are shown in \cref{fig:supp_multires_rendering}. The model was trained as described in the main text with 4 resolutions $r^{(i)} \in \{1, 0.1, 0.05, 0.01\}$. The results show that, indeed, the coarser resolutions capture coarse and more global texture features that are then refined by including the finer resolutions in the feature gathering.

\subsection{Additional Results for the Texture Reconstruction}
\label{sec:supp_add_results_texture_recon}
\begin{figure}[ht]
    \centering

    \begin{subfigure}{0.18\textwidth}
        \includegraphics[width=\linewidth]{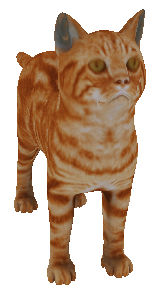}
    \end{subfigure}
    \hfill
    \begin{subfigure}{0.18\textwidth}
        \includegraphics[width=\linewidth]{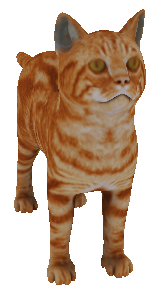}
    \end{subfigure}
    \hfill
    \begin{subfigure}{0.18\textwidth}
        \includegraphics[width=\linewidth]{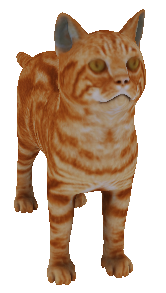}
    \end{subfigure}
    \hfill
    \begin{subfigure}{0.18\textwidth}
        \includegraphics[width=\linewidth]{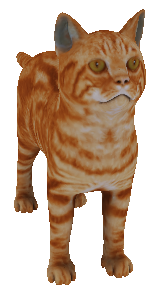}
    \end{subfigure}
    \hfill
    \begin{subfigure}{0.18\textwidth}
        \includegraphics[width=\linewidth]{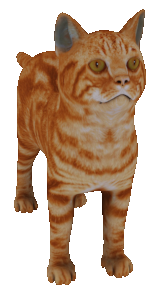}
    \end{subfigure}

    \vspace{1pt} %

    \begin{subfigure}{0.18\textwidth}
        \includegraphics[width=\linewidth]{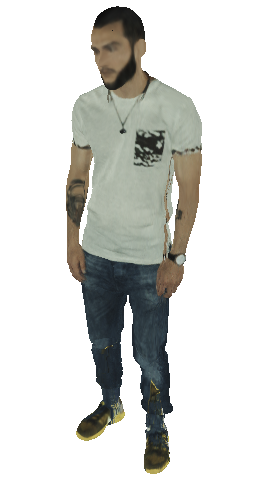}
        \caption{NeuTex}
    \end{subfigure}
    \hfill
    \begin{subfigure}{0.18\textwidth}
        \includegraphics[width=\linewidth]{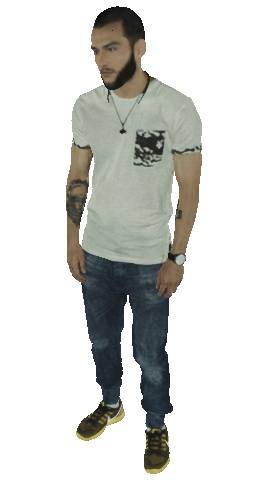}
        \caption{TF+RFF}
    \end{subfigure}
    \hfill
    \begin{subfigure}{0.18\textwidth}
        \includegraphics[width=\linewidth]{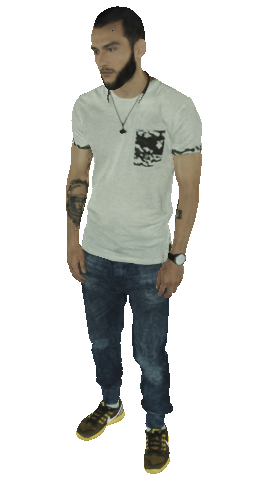}
        \caption{INF}
    \end{subfigure}
    \hfill
    \begin{subfigure}{0.18\textwidth}
        \includegraphics[width=\linewidth]{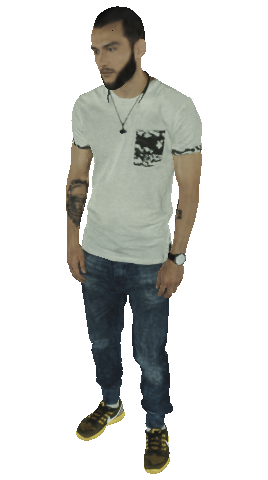}
    \caption{Ours}
    \end{subfigure}
    \hfill
    \begin{subfigure}{0.18\textwidth}
        \includegraphics[width=\linewidth]{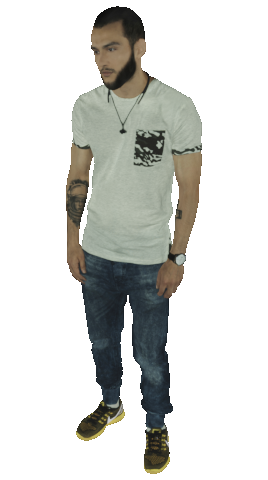}
        \caption{GT}
    \end{subfigure}

    \caption{Further qualitative comparisons on unseen views to the baseline methods \cite{Xiang21NeuTex, Oechsle19TextureFields, Koestler22IntrinsicNeuralFields}. We produce results that are on par with the state-of-the-art while providing our notable speedup. 
    Note that we use 
    $d=10$ for the latent codes of the cat object in this figure.
    }
    \label{fig:supp_tex_recon_qual}
\end{figure}

\begin{figure}[t]
    \centering

    \begin{subfigure}{0.19\textwidth}
        \includegraphics[
        width=\linewidth,
        trim={6cm 2cm 6cm 0},clip
    ]{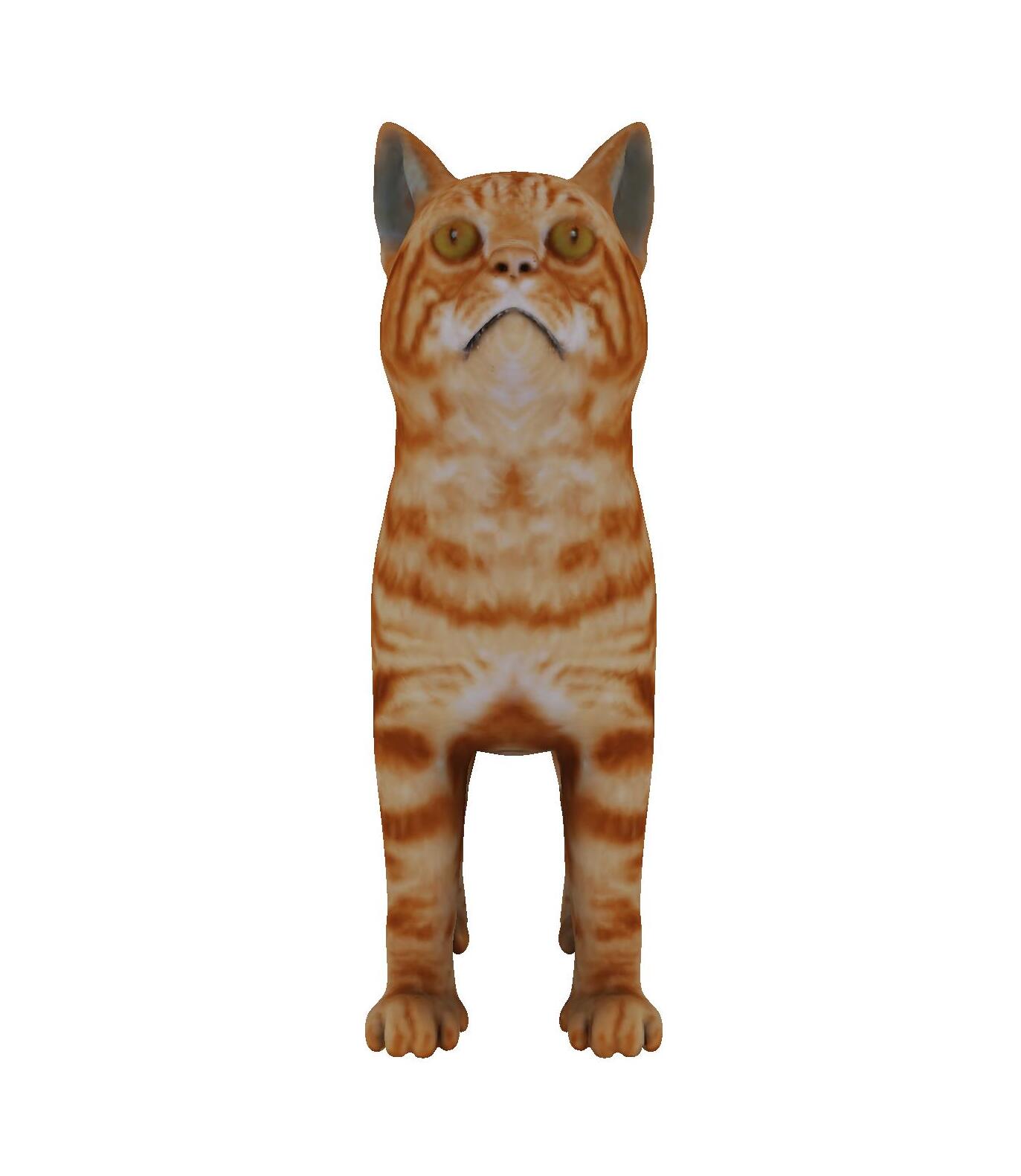}
    \end{subfigure}
    \hfill
    \begin{subfigure}{0.19\textwidth}
        \includegraphics[
        width=\linewidth,
        trim={10cm 0 2cm 2cm},clip
    ]{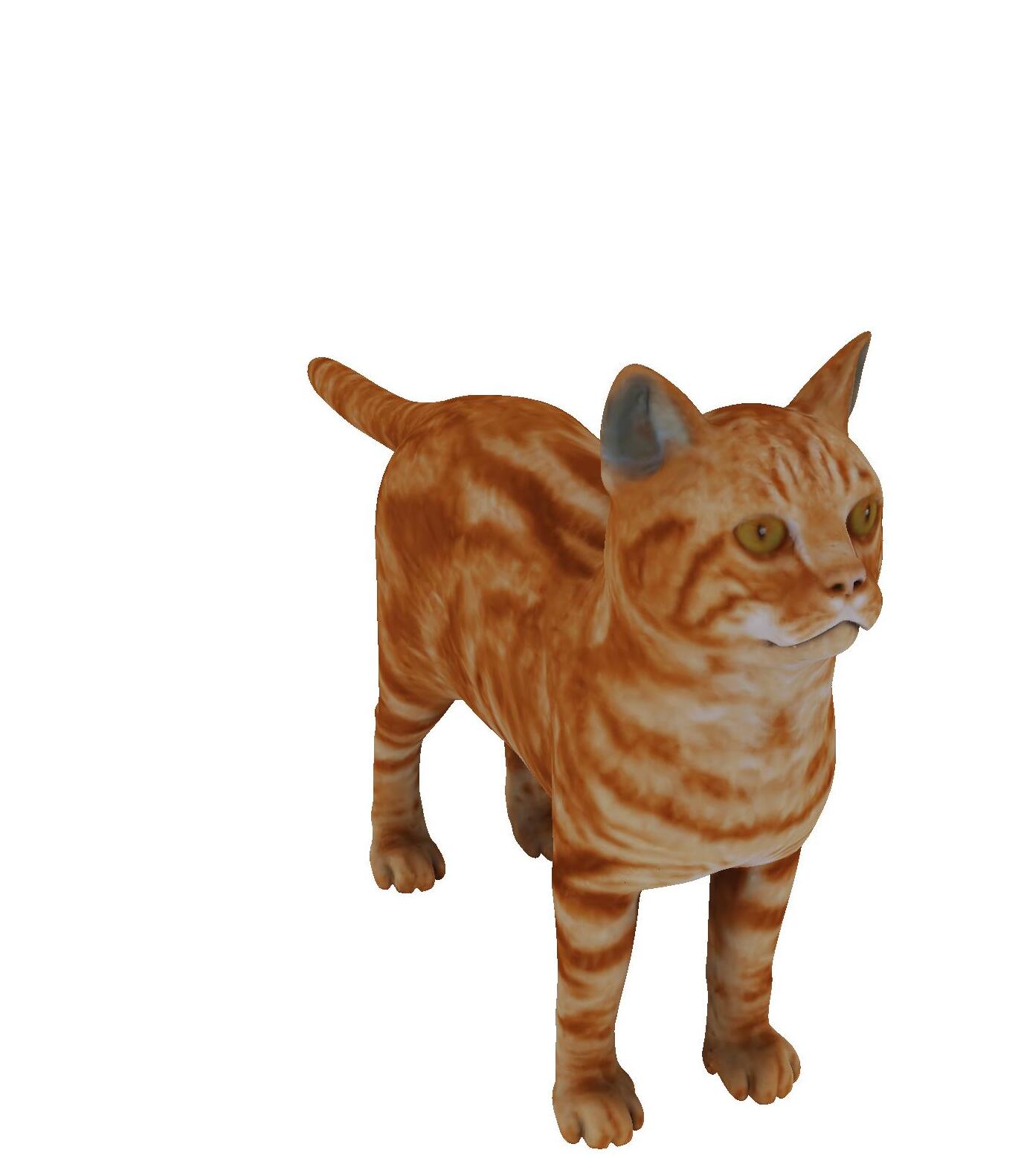}
    \end{subfigure}
    \hfill
    \begin{subfigure}{0.19\textwidth}
        \includegraphics[
        width=\linewidth,
        trim={6cm 0 6cm 2cm},clip
    ]{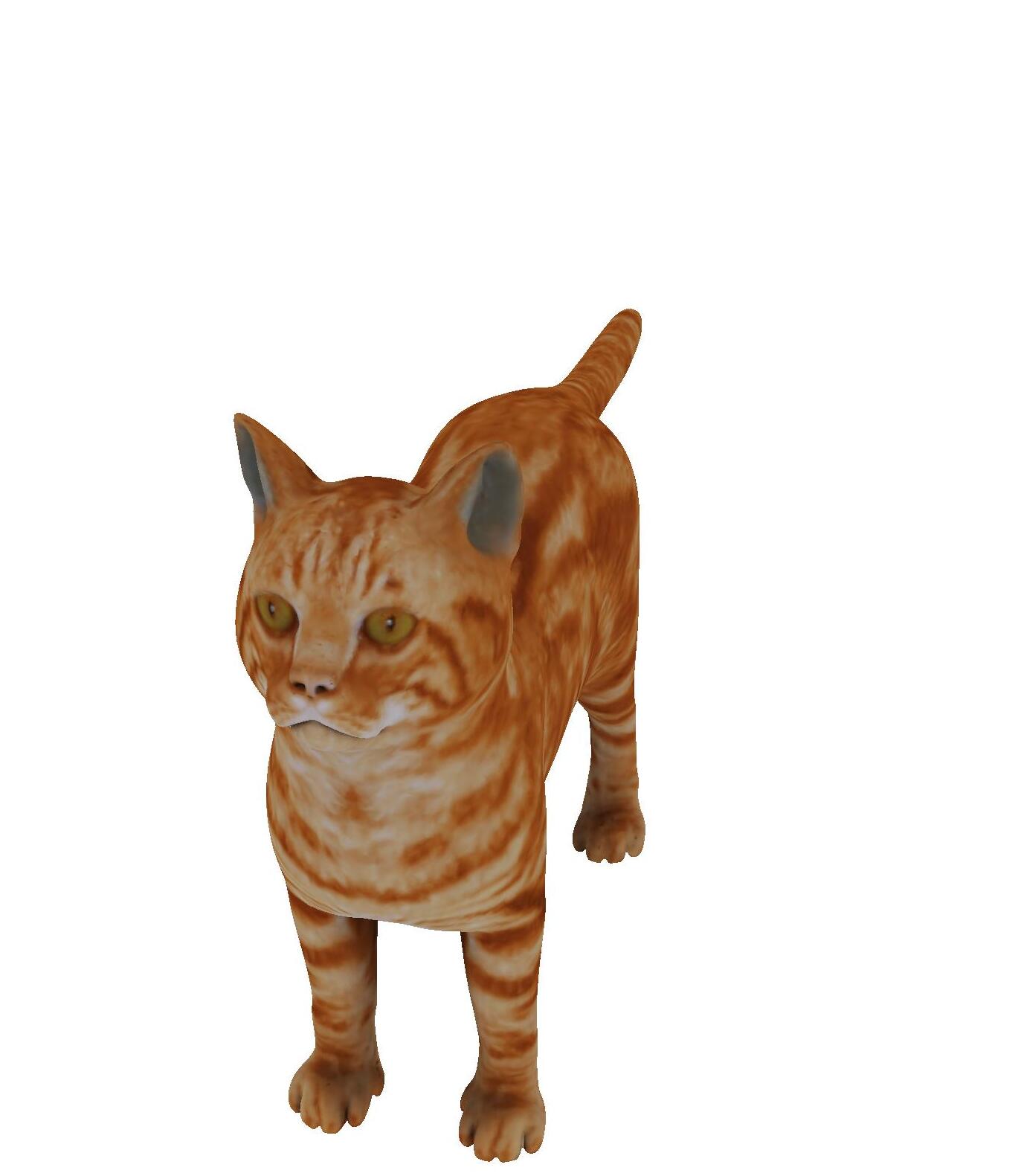}
    \end{subfigure}
    \hfill
    \begin{subfigure}{0.19\textwidth}
        \includegraphics[
        width=\linewidth,
        trim={3cm 2cm 8cm 0},clip
    ]{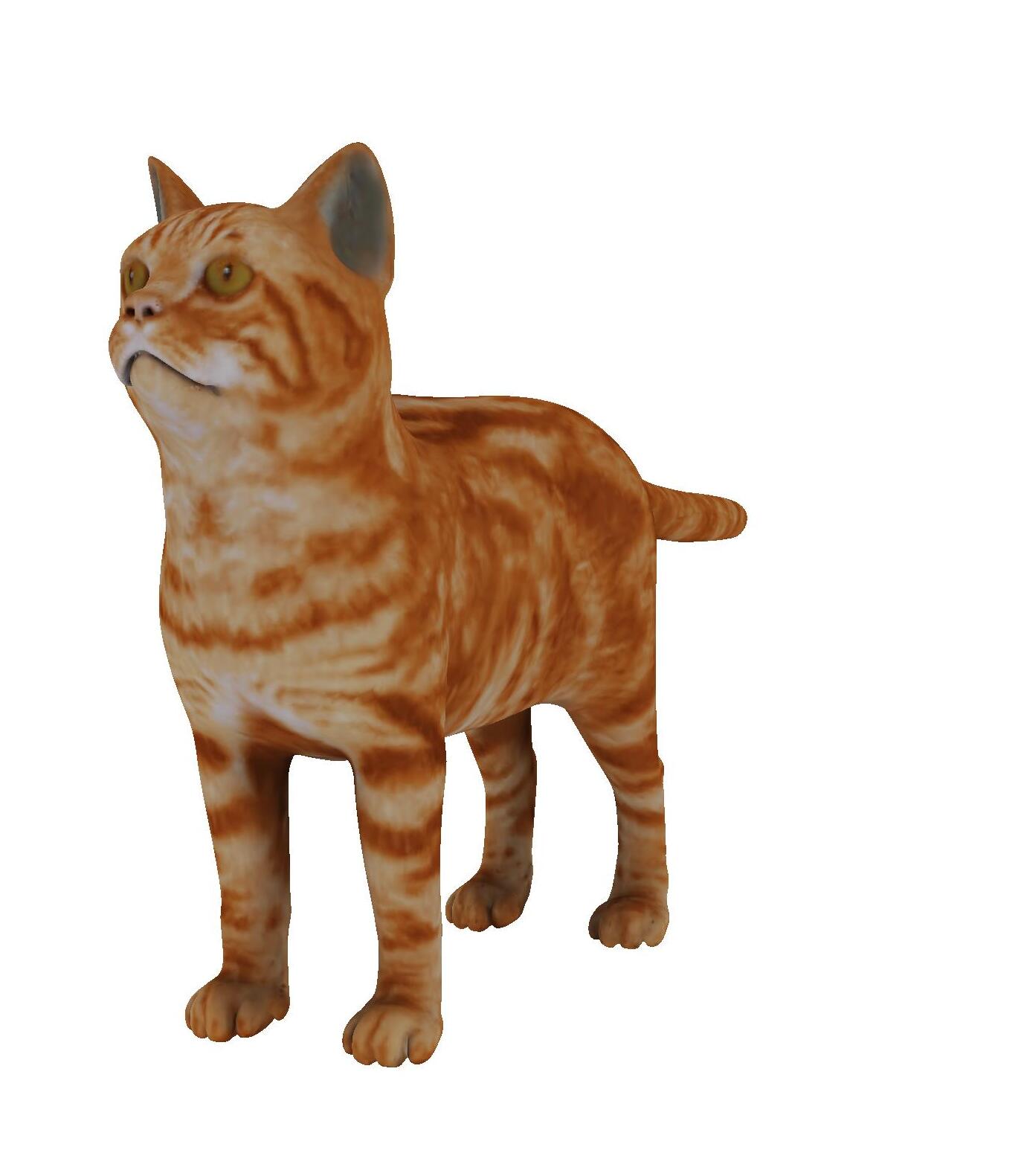}
    \end{subfigure}
    \hfill
    \begin{subfigure}{0.19\textwidth}
        \includegraphics[
        width=\linewidth,
        trim={8cm 2cm 3cm 0},clip
    ]{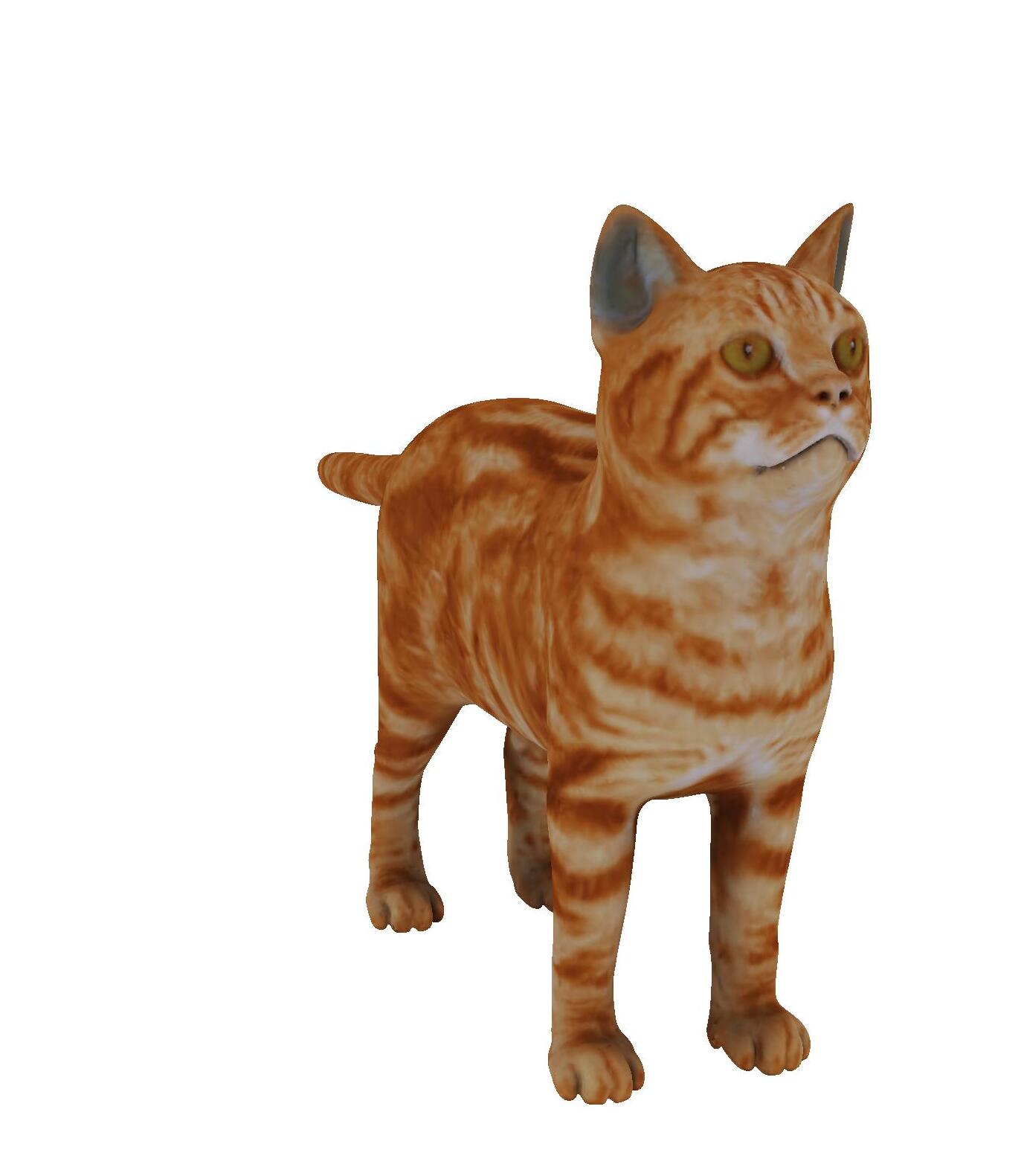}
    \end{subfigure}

    \vspace{15pt} %

    \begin{subfigure}{0.19\textwidth}
        \includegraphics[width=\linewidth]{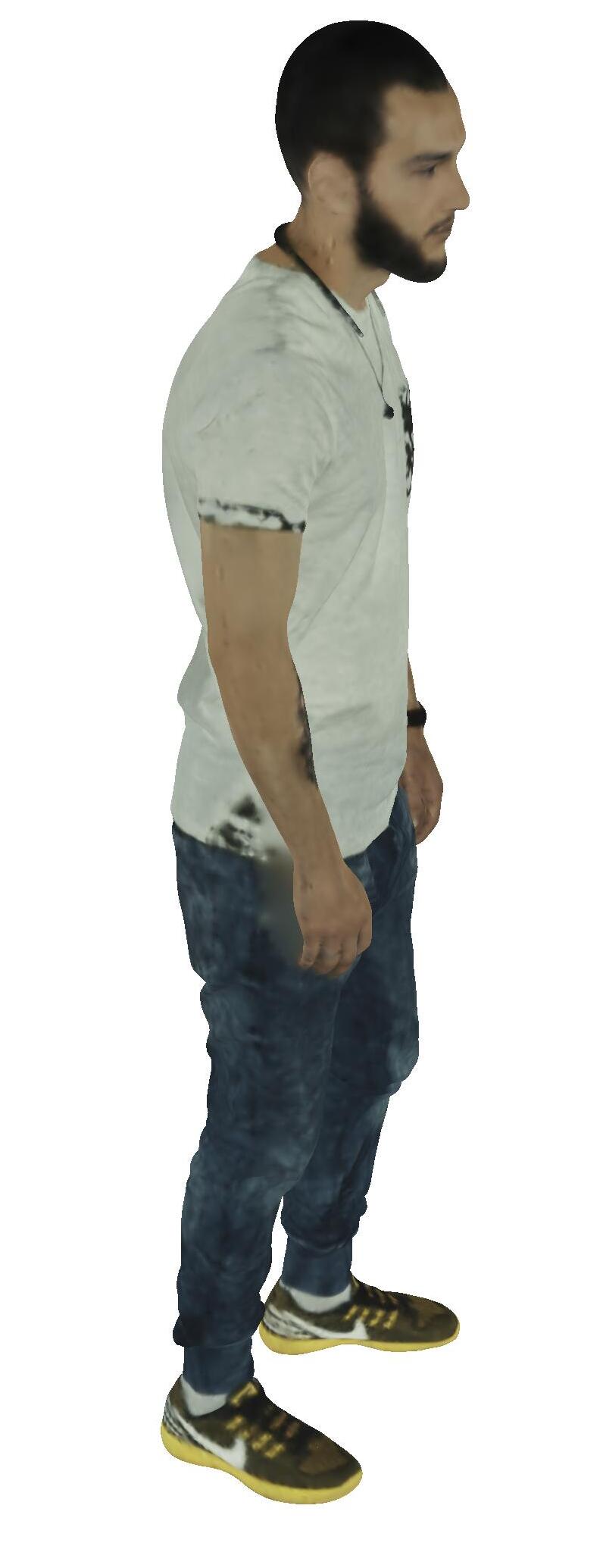}
    \end{subfigure}
    \hfill
    \begin{subfigure}{0.19\textwidth}
        \includegraphics[width=\linewidth]{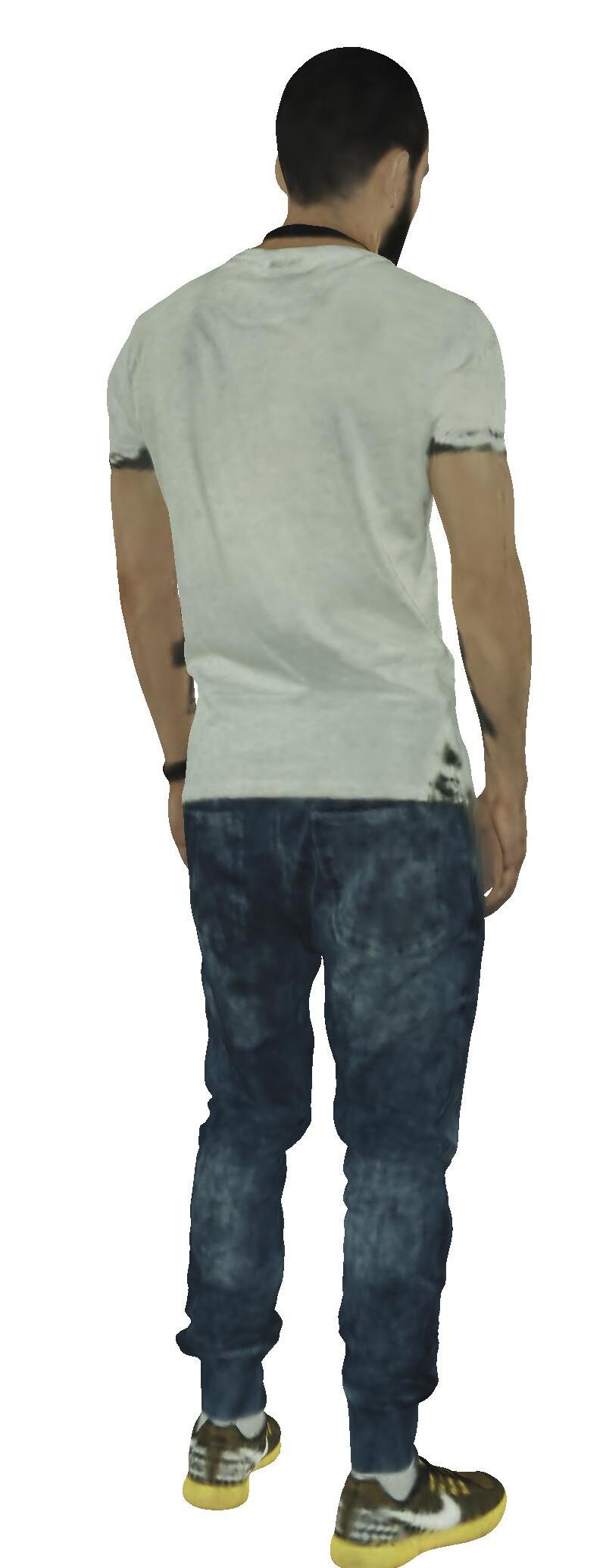}
    \end{subfigure}
    \hfill
    \begin{subfigure}{0.19\textwidth}
        \includegraphics[width=\linewidth]{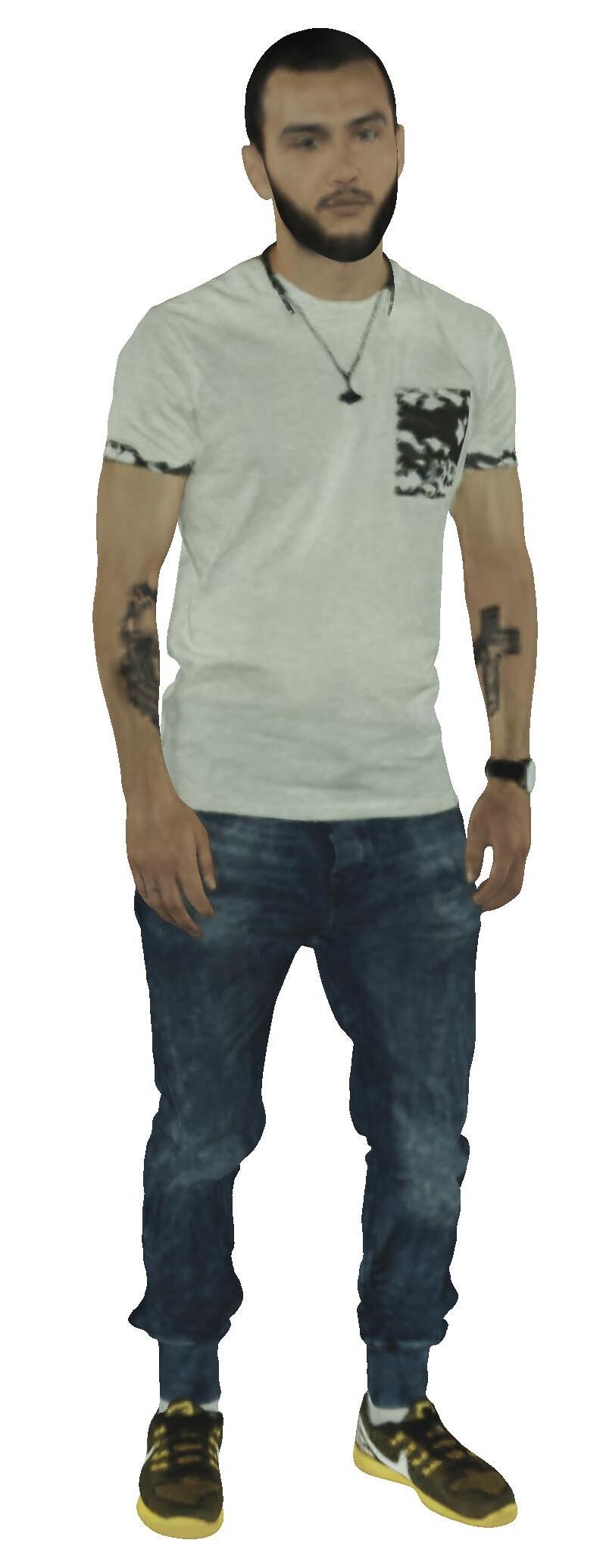}
    \end{subfigure}
    \hfill
    \begin{subfigure}{0.19\textwidth}
        \includegraphics[width=\linewidth]{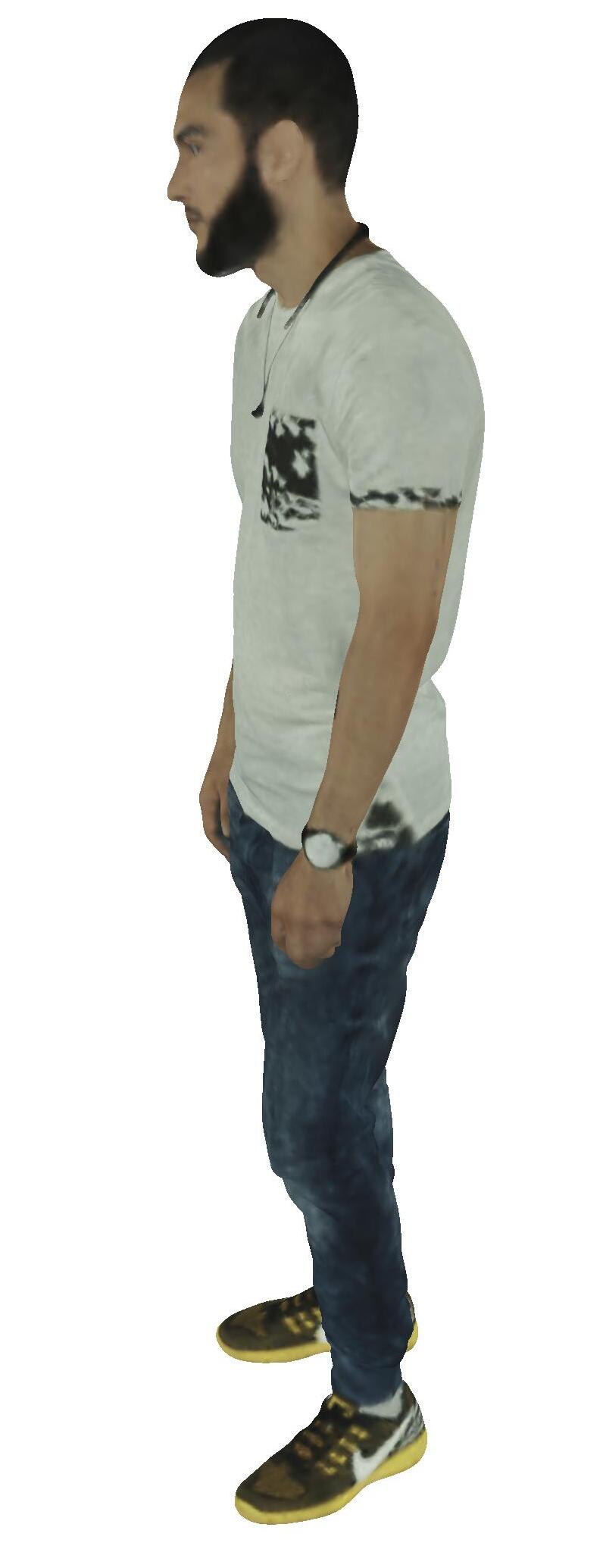}
    \end{subfigure}
    \hfill
    \begin{subfigure}{0.19\textwidth}
        \includegraphics[width=\linewidth]{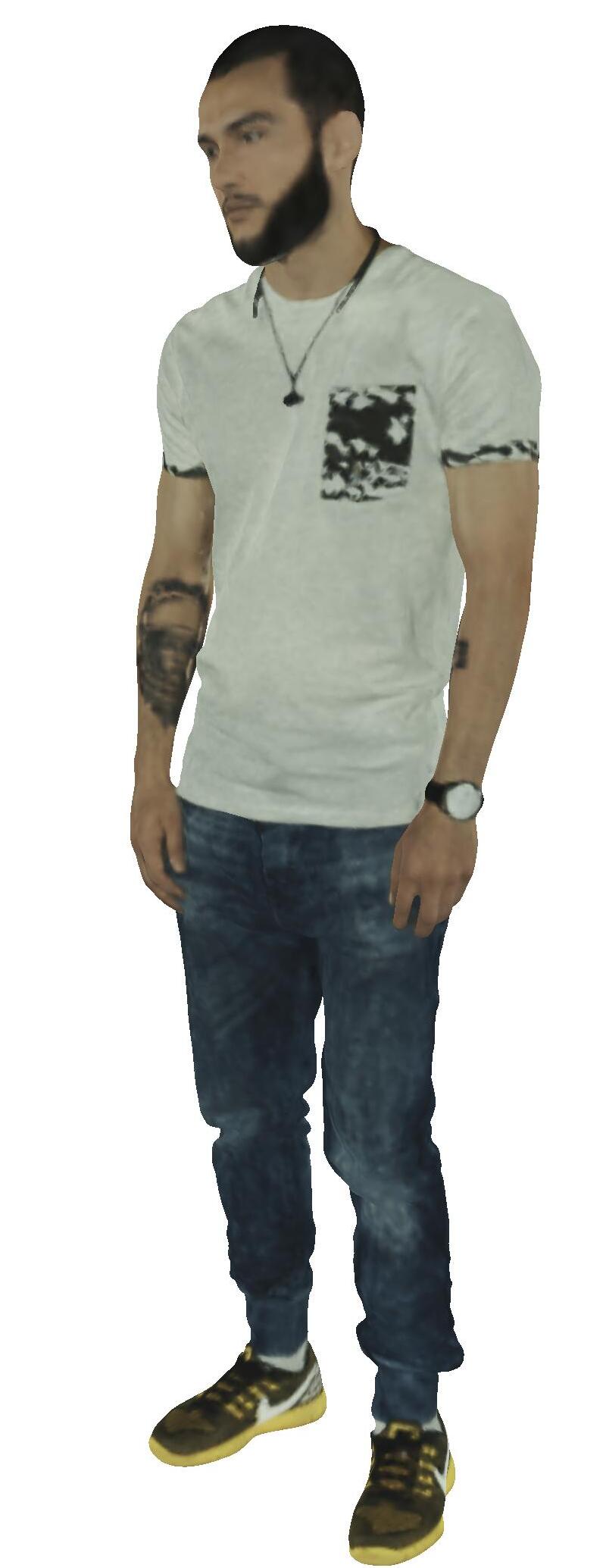}
    \end{subfigure}

    \caption{Further renderings for unseen views for the cat and human object using MeshFeat. We see, that our method enables good reconstruction on all parts of the mesh.}
    \label{fig:supp_renderings}
\end{figure}

We show additional qualitative comparisons between the methods in \cref{fig:supp_tex_recon_qual}. Our method yields high-quality reconstruction results on par with the other methods while admitting the significantly faster inference described in the main paper. In \cref{fig:supp_renderings}, we present additional views for our method, showing that we consistently achieve a good reconstruction on all mesh parts.

\subsection{Further analysis on the choice of hyperparameters}
\label{sec:supp_add_analysis_hyperparams}

\subsubsection{MLP Parameters}

One of the main reasons for our significant speedup is the very shallow MLP with 2 hidden layers with a dimension of 32. 
Since the latent features contain the spatial information of the texture, the MLP only needs to decode them into RGB values.
In \cref{tab:mlp_params}, we analyze the effect of reducing the MLP size even further. However, reducing the hidden dimension or the number of layers yields diminishing returns.

\begin{table}[ht]
\centering
\begin{tabular}{|c | c || c c c|}
\hline
 Hidden Layers &  Hidden dimension &  PSNR$\uparrow$ &  DSSIM $\downarrow$ &  LPIPS $\downarrow$ \\ \hline
2 & 16 & 32.31 & 0.209 & 0.417 \\ 
2 & 32 & \textbf{32.51} & \textbf{0.202} & 0.400 \\
1 & 32 & 32.39 & 0.205 & 0.402 \\
1 & 64 & 32.42 & 0.206 & \textbf{0.395} \\  
1 & 128 & 32.39 & 0.210 & 0.398 \\ \hline
\end{tabular}
\vspace{0.3cm}
\caption{
Reconstruction quality with different numbers of layers and hidden dimensions.
We observe a slight decrease in reconstruction quality after decreasing 
the hidden dimension or the number of layers even further. For a single layer, the results are slightly worse, even for a significantly increased hidden dimension.
}
\label{tab:mlp_params}
\end{table}

\subsubsection{Encoding dimensions}
\begin{table}[ht]
\centering
\begin{tabular}{|c | c c c|}
\hline
 Latent code dim ($d$) & PSNR $\uparrow$ & DSSIM $\downarrow$ & LPIPS $\downarrow$ \\ \hline
1 & 29.30 & 0.3209 & 0.8058 \\ 
2 & 32.28 & 0.2134 & 0.4182 \\ 
3 & 32.21 & 0.2166 & 0.4095 \\ 
4 & 32.51 & 0.2019 & 0.3962 \\ 
5 & 32.42 & \textbf{0.2001} & 0.4053 \\ 
6 & \textbf{32.65} & 0.2003 & 0.3984 \\ 
7 & 32.43 & 0.2022 & \textbf{0.3342} \\ 
8 & 32.46 & 0.2003 & 0.4051 \\ 
9 & 32.51 & 0.2014 & 0.4045 \\ 
10 & 32.43 & 0.2043 & 0.4035 \\ 
11 & 32.43 & 0.2001 & 0.4113 \\ 
12 & 32.48 & 0.2060 & 0.4015 \\ 
13 & 32.48 & 0.2039 & 0.4191 \\ 
14 & 32.43 & 0.2023 & 0.4050 \\ 
15 & 32.47 & 0.2036 & 0.3991 \\ \hline
\end{tabular}
\vspace{0.3cm}
\caption{
Influence of the encoding dimensions $d$ for texture reconstruction on the human object. 
While the reconstruction quality is fairly similar over a large range of dimensions, we observe a slight decrease for increasing values of $d$, which might indicate the tendency of overfitting. Also, we see a significant drop at the low end of the table, which shows that a minimum number of encoding dimensions is required to achieve high-quality reconstructions.
}
\label{tab:enc_dims}
\end{table}
The dimension $d$ of our features is the main factor for the model size, and therefore a small dimension is desirable. However, the model size needs to be balanced with the reconstruction quality.
\cref{tab:enc_dims} shows the reconstruction quality over a large range of encoding dimensions. We observe only a slight influence of the encoding dimension on the reconstruction quality, with a slight indication of overfitting for increasing the dimension. For very low dimensions, the results show a significant drop, indicating that the model is unable to reconstruct the texture faithfully, given too few features to store the information.

\subsubsection{Resolution configurations}
\begin{table}[ht]
\centering
\begin{tabular}{|l|l|l|l|}
\hline
 Used Resolutions & PSNR $\uparrow$ & DSSIM $\downarrow$ & LPIPS $\downarrow$ \\ \hline
\{1, 0.1, 0.05, 0.01\} & \textbf{32.51} & \textbf{0.2019} & 0.3962 \\ 
\{1, 0.1, 0.01\} & 32.44 & 0.2024 & 0.4010 \\
\{1, 0.5, 0.25, 0.125\} & 32.35 & 0.2118 & 0.3899 \\ 
\{1, 0.25, 0.0625, 0.0625\} & 32.38 & 0.2061 & \textbf{0.3874} \\ 
\{0.75, 0.25, 0.075, 0.025\} & 31.51 & 0.2728 & 0.4811 \\
\{0.9, 0.12, 0.05, 0.01\} & 31.99 & 0.2301 & 0.4303 \\ \hline
\end{tabular}
\vspace{0.3cm}
\caption{
Influence of the choice of  resolutions $r^{(i)}$ on the reconstruction quality.
We observe a fairly stable reconstruction quality for different resolution combinations, as long as the finest resolution $r^{(1)}=1$ is included. 
}
\label{tab:resolution_configs}
\end{table}
Our method leverages a mesh simplification algorithm \cite{GarlandH97SurfaceSimplification, Garland98SimplifyingSurfacesWithColorAndTexture} to obtain different mesh resolutions, which are then used for the multi-resolution approach. It is apparent through \cref{tab:resolution_configs} that the method is robust to different resolution configurations
as long as the finest resolution is the original resolution, \ie $r^{(1)}=1$.

\subsection{Absolute inference speed}
To gain more insight into the inference speedup, we provide absolute values for inference speed in \cref{tab:abs_speed}. 
\begin{table*}[ht]
\centering
\begin{tabular}{|l || l |l |}
\hline
 Method & human & cat \\ \hline
 Neutex & 14.738ms & 14.453ms \\
 TF+RFF & 7.120ms & 7.090ms \\ 
 INF & 5.000ms & 4.994ms \\ 
 Ours d=4 & 1.144ms & 1.021ms \\ 
 Ours d=10 & 1.409ms & 1.381ms \\ \hline
\end{tabular}
\vspace{0.3cm}
 \caption{
 Absolute inference speed in milliseconds for texture representation on the human object.
 Our time measurement includes a GPU warmup over 10 steps. The reported absolute inference time is the time taken for a forward pass for each method averaged over 300 repetitions.
 }
    \label{tab:abs_speed}
\end{table*}

\subsection{Qualitative comparison to a single resolution approach}
\begin{figure}[ht]
    \centering

    \scriptsize
  
    \newcommand{\mywidthx}{0.22\textwidth}  %
    \newcommand{\myheightx}{0.3\textwidth}  %
  
    \newcolumntype{X}{ >{\centering\arraybackslash} m{\mywidthx} } %

    \def\arraystretch{1} %
  \begin{tabular}{XXXX}
    \includegraphics[width=\mywidthx]{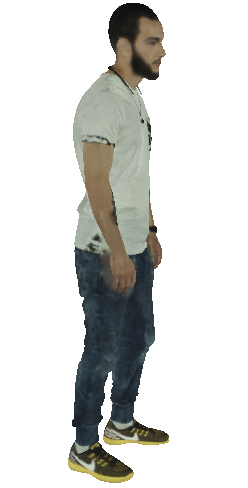} &
    \includegraphics[width=\mywidthx]{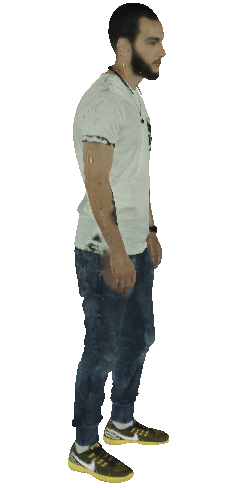} &
    \includegraphics[width=\mywidthx]{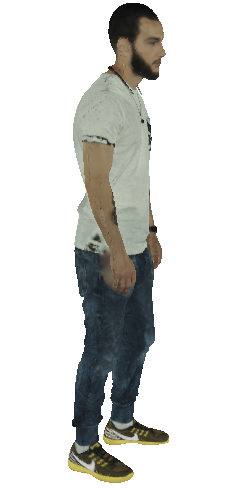} &
    \includegraphics[width=\mywidthx]{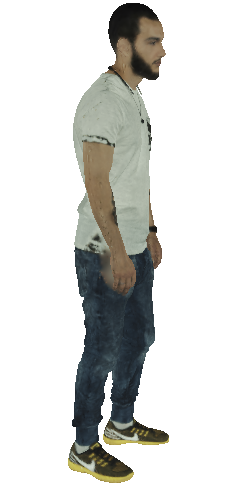} \\

    Multiresolution ($d=4$) &
    Single resolution ($d=4$) &
    Single resolution ($d=6$) &
    Single resolution ($d=10$) \\

  \end{tabular}

    \caption{
    Qualitative comparison between the multi-resolution approach and reconstructions using features over only a single resolution (the original mesh resolution). The single-resolution approach leads to more noisy reconstructions despite the increased dimension ($d$) of the feature vectors. 
    }
    \label{fig:supp_multires_vs_singleres}
\end{figure}

\iftoggle{arxiv}{\cref{sec:ablation}}{Sec. 4.2}
in the main text demonstrates how a single-resolution setup struggles to achieve a good reconstruction. 
\cref{fig:supp_multires_vs_singleres} shows that the results of the single-resolution show noisy areas despite using the regularizer. This might indicate that the multi-resolution enables the regularizer to act more efficiently since it increases the area of influence through the coarse resolutions.

\subsection{Additional Results for the BRDF Estimation}
\label{sec:supp_add_results_brdf}
\begin{figure}[t]
    \centering
    \scriptsize
  
    \newcommand{\mywidthx}{0.22\textwidth}  %
    \newcommand{\myheightx}{0.3\textwidth}  %
    \newcommand{\mywidtht}{0.045\textwidth}  %
  
    \newcolumntype{X}{ >{\centering\arraybackslash} m{\mywidthx} } %
    \newcolumntype{T}{ >{\centering\arraybackslash} m{\mywidtht} } %

    \def\arraystretch{1} %
  \begin{tabular}{TXXXX}
  \rotatebox{90}{\parbox{3cm}{\normalsize\centering bear}} &
  \includegraphics[width=\mywidthx, height=\myheightx, keepaspectratio]{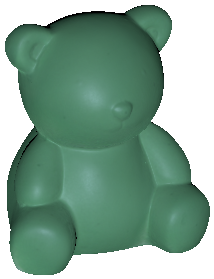} &
  \includegraphics[width=\mywidthx, height=\myheightx, keepaspectratio]{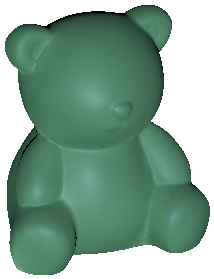} & 
  \includegraphics[width=\mywidthx, height=\myheightx, keepaspectratio]{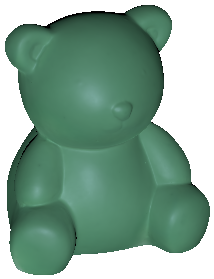} &
  \includegraphics[width=\mywidthx, height=\myheightx, keepaspectratio]{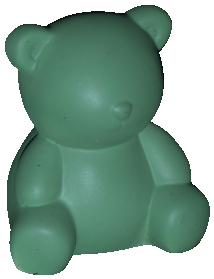} \\
  \rotatebox{90}{\parbox{3cm}{\normalsize\centering buddha}} &
  \includegraphics[width=\mywidthx, height=\myheightx, keepaspectratio]{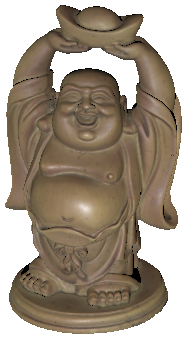} &
  \includegraphics[width=\mywidthx, height=\myheightx, keepaspectratio]{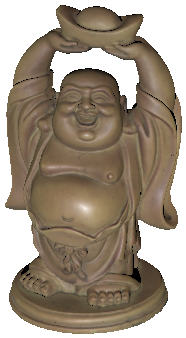} & 
  \includegraphics[width=\mywidthx, height=\myheightx, keepaspectratio]{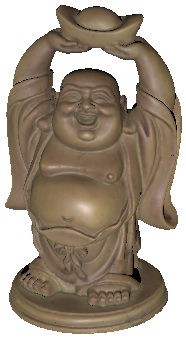} &
  \includegraphics[width=\mywidthx, height=\myheightx, keepaspectratio]{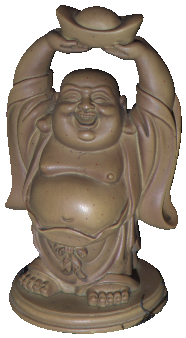} \\
  \rotatebox{90}{\parbox{3cm}{\normalsize\centering cow}} &
  \includegraphics[width=\mywidthx, height=\myheightx, keepaspectratio]{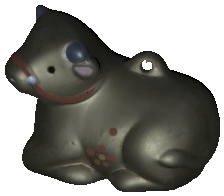} &
  \includegraphics[width=\mywidthx, height=\myheightx, keepaspectratio]{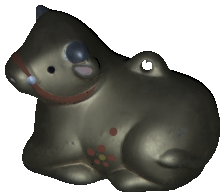} & 
  \includegraphics[width=\mywidthx, height=\myheightx, keepaspectratio]{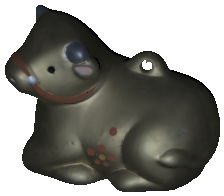} &
  \includegraphics[width=\mywidthx, height=\myheightx, keepaspectratio]{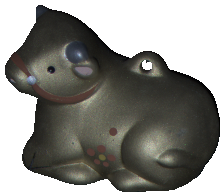} \\
  \rotatebox{90}{\parbox{3cm}{\normalsize\centering pot2}} &
  \includegraphics[width=\mywidthx, height=\myheightx, keepaspectratio]{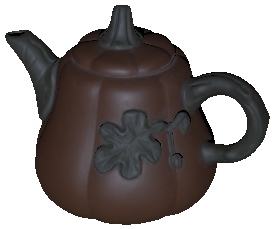} &
  \includegraphics[width=\mywidthx, height=\myheightx, keepaspectratio]{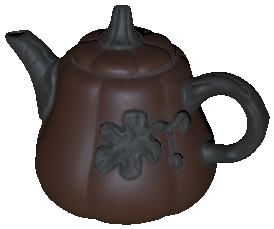} & 
  \includegraphics[width=\mywidthx, height=\myheightx, keepaspectratio]{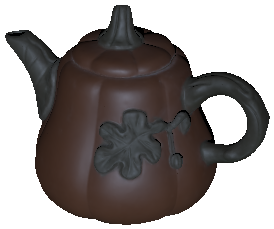} &
  \includegraphics[width=\mywidthx, height=\myheightx, keepaspectratio]{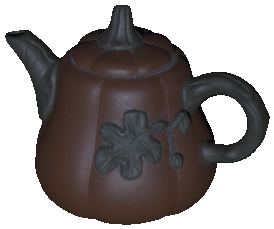} \\
  \rotatebox{90}{\parbox{3cm}{\normalsize\centering reading}} &
  \includegraphics[width=\mywidthx, height=\myheightx, keepaspectratio]{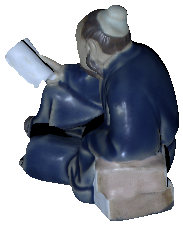} &
  \includegraphics[width=\mywidthx, height=\myheightx, keepaspectratio]{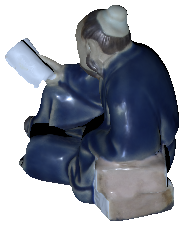} & 
  \includegraphics[width=\mywidthx, height=\myheightx, keepaspectratio]{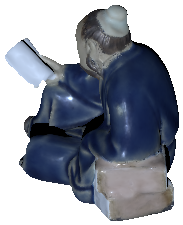} &
  \includegraphics[width=\mywidthx, height=\myheightx, keepaspectratio]{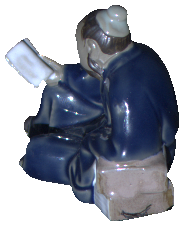} \\
  & \textbf{(a)} TF+RFF
  & \textbf{(b)} INF
  & \textbf{(c)} Ours
  & \textbf{(d)} GT \\

  \end{tabular}

    \caption{Qualitative Results for the BRDF reconstruction for all objects of the DiLiGenT-MV dataset. The results of our method are practically indistinguishable from the results of the other methods.
    }
    \label{fig:supp_brdf_recon_qualitative}
\end{figure}

For completeness, we show the quantitative results for all 5 objects of the DiLiGenT-MV dataset individually in \cref{tab:supp_brdf_recon_quant}. The results confirm, that we achieve results that are consistently of comparable reconstruction quality to the other methods while achieving a significant inference speed-up. Qualitative results in the form of a single view per object are presented in \cref{fig:supp_brdf_recon_qualitative}. We see that for all objects in the dataset, the results of our method are hardly distinguishable from those of the other methods.

    \begin{table*}[t]  %
    \centering  %
    \begin{tabular}{ | l || l | l l l l l |} 
    \hline
    Object & Method & PSNR $\uparrow$ & DSSIM $\downarrow$ & LPIPS $\downarrow$ & \# Params. 
    $\downarrow$ & Speedup $\uparrow$  \\	\hline
\hline 

\multirow{1}{5em}{bear} 
& TF+RFF & 43.68 & 0.4826 & \textbf{0.9227} \goldmedal & \textbf{332k} \goldmedal & 1.00x \\ 
& INF & \textbf{43.78} \goldmedal & \textbf{0.4772} \goldmedal & 0.9730 \silvermedal & 204931k & 1.08x \silvermedal \\ 
& Ours & 43.73 \silvermedal & 0.4778 \silvermedal & 0.9913 & 930k \silvermedal & \textbf{7.78x} \goldmedal \\ 
\hline\hline
\multirow{1}{5em}{buddha} 
& TF+RFF & 37.03 \silvermedal & \textbf{1.0785} \goldmedal & \textbf{2.0095} \goldmedal & \textbf{332k} \goldmedal & 1.00x \\ 
& INF & 37.02 & 1.0904 & 2.0393 \silvermedal & 204929k & 1.08x \silvermedal \\ 
& Ours & \textbf{37.04} \goldmedal & 1.0862 \silvermedal & 2.0569 & 930k \silvermedal & \textbf{7.35x} \goldmedal \\ 
\hline\hline
\multirow{1}{5em}{cow} 
& TF+RFF & 47.22 & 0.3318 & \textbf{1.1384} \goldmedal & \textbf{332k} \goldmedal & 1.00x \\ 
& INF & \textbf{47.31} \goldmedal & \textbf{0.3299} \goldmedal & 1.2184 \silvermedal & 204931k & 1.08x \silvermedal \\ 
& Ours & 47.29 \silvermedal & 0.3300 \silvermedal & 1.4572 & 930k \silvermedal & \textbf{7.62x} \goldmedal \\ 
\hline\hline
\multirow{1}{5em}{pot2} 
& TF+RFF & 46.69 & 0.4581 & 0.9326 \silvermedal & \textbf{332k} \goldmedal & 1.00x \\ 
& INF & \textbf{46.81} \goldmedal & \textbf{0.4485} \goldmedal & \textbf{0.9317} \goldmedal & 204927k & 1.08x \silvermedal \\ 
& Ours & 46.80 \silvermedal & 0.4518 \silvermedal & 0.9802 & 930k \silvermedal & \textbf{7.72x} \goldmedal \\ 
\hline\hline
\multirow{1}{5em}{reading} 
& TF+RFF & 36.02 \silvermedal & 1.0079 & 2.5034 \silvermedal & \textbf{332k} \goldmedal & 1.00x \\ 
& INF & \textbf{36.14} \goldmedal & \textbf{0.9832} \goldmedal & \textbf{2.4690} \goldmedal & 204929k & 1.08x \silvermedal \\ 
& Ours & 36.02 & 1.0041 \silvermedal & 2.5352 & 930k \silvermedal & \textbf{7.44x} \goldmedal \\ 
\hline

    \end{tabular}
    \vspace{0.3cm}

    \caption{Quantitative results of the BRDF reconstruction for all five objects of the DiLiGenT-MV dataset. Note that DSSIM and LPIPS are scaled by 100. The results show, that our method yields reconstruction quality that is on par with the other methods while achieving a significant speed-up for the inference. 
    }
    \label{tab:supp_brdf_recon_quant}
    \end{table*}

\end{document}